\documentclass[journal]{IEEEtranTIE}
\usepackage{graphicx}
\usepackage{cite}
\usepackage{picinpar}
\usepackage{amsmath}
\usepackage{url}
\usepackage{flushend}
\usepackage[latin1]{inputenc}
\usepackage{colortbl}
\usepackage{soul}
\usepackage{multirow}
\usepackage{pifont}
\usepackage{color}
\usepackage{alltt}
\usepackage[hidelinks]{hyperref}
\usepackage{enumerate}
\usepackage{siunitx}
\usepackage{breakurl}
\usepackage{epstopdf}
\usepackage{pbox}

\usepackage{subfigure}
\usepackage{algorithm}
\usepackage{algorithmic}
\usepackage{tablefootnote}
\usepackage{mathrsfs}
\usepackage{array}
\usepackage{caption}
\usepackage{booktabs}
\usepackage{threeparttable}

\graphicspath{{figure/}}
\newcommand{\tabincell}[2]{\begin{tabular}{@{}#1@{}}#2\end{tabular}}

\begin{document}
	\title{	Aggressive Quadrotor Flight Using Curiosity-Driven Reinforcement Learning}
	
	\author{
		\vskip 1em
		{		 
		Qiyu Sun,
		Jinbao Fang, 
		Wei Xing Zheng, \emph{Fellow, IEEE,}
		and Yang Tang, \emph{Senior Member, IEEE}
		}
	
		\thanks{
			
			{						
			Q. Sun, J. Fang,  and Y. Tang are with the Key Laboratory of Smart Manufacturing in Energy Chemical Process Ministry of Education, East China University of Science and Technology, Shanghai, 200237, China (e-mail: yangtang@ecust.edu.cn (Y. Tang)).
			
			W. X. Zheng is with the School of Computer, Data and Mathematical Sciences,
			Western Sydney University, Sydney, NSW 2751, Australia (e-mail:
			w.zheng@westernsydney.edu.au).
			
			}
		}
	}
	
	\maketitle
	
	\begin{abstract}
		The ability to perform aggressive movements, which are called aggressive flights, is important for quadrotors during navigation. However, aggressive quadrotor flights are still a great challenge to practical applications.  The existing solutions to aggressive flights heavily rely on a predefined trajectory, which is a time-consuming preprocessing step. To avoid such path planning, we propose a curiosity-driven reinforcement learning method for aggressive flight missions and a similarity-based curiosity module is introduced to speed up the training procedure. A branch structure exploration (BSE) strategy  is also applied to guarantee the robustness of the policy and to ensure the policy trained in simulations can be performed in real-world experiments directly.  The experimental results in simulations demonstrate that our  reinforcement learning algorithm performs well in aggressive flight tasks,  speeds up the convergence process and improves the robustness of the policy. Besides, our algorithm shows a satisfactory simulated to real transferability and performs well in  real-world experiments. 
	\end{abstract}
	
	\begin{IEEEkeywords}
		UAVs, Aggressive Flight, Reinforcement Learning
	\end{IEEEkeywords}
	
	
	\definecolor{limegreen}{rgb}{0.2, 0.8, 0.2}
	\definecolor{forestgreen}{rgb}{0.13, 0.55, 0.13}
	\definecolor{greenhtml}{rgb}{0.0, 0.5, 0.0}
	
	\section{Introduction}
	
	Unmanned aerial vehicles (UAVs) are well-performing platforms for many tasks, such as exploration~\cite{exploration}, rescue assistance~\cite{rescue} and surveillance~\cite{surveillance}. In particular, UAVs are suitable for executing  tasks such as agilely moving  and avoiding obstacles with high linear and angular  speed, which are called  aggressive flights. The ability of aggressive flight is significant for UAVs, especially in the situations such as  entering a damaged building with cluttered obstacles to find trapped persons~\cite{rescue}. There are several representative tasks in aggressive flight problems, such as flying through narrow windows and slalom path scenes with a high speed~\cite{KumarAggressive}. In these missions, UAVs are confronted with the challenges of continuously performing precise aggressive movements and stabilizing themselves in aggressive states.

	
	
	Previous works~\cite{KumarAggressive,ActVision} have demonstrated the possibility of navigating UAVs in aggressive flight missions. 	
	{However, a time-consuming trajectory planning procedure is commonly required to provide an aggressive trajectory before the navigation of UAVs in these methods. 
	As this pre-required aggressive trajectory planning may take considerable time in some unstructured environments~\cite{LiuTrajPlan}, it restricts applications with a tight time constraint such as rescue assistance. Thus,  an end-to-end navigation strategy that is able to directly use current state information to generate a control command is of great importance and is considered to be a superior solution in the tasks expecting rapid reactions. 
One of the promising methods to address this is reinforcement learning, through which UAVs can be navigated  without the trajectory planning procedure~\cite{RL-MPC} and which has been widely used in the control and navigation of agents~\cite{TIE1, TIE2,TIE3, TIE4}. }

	\begin{figure}[t]
	\setlength{\abovecaptionskip}{0.cm}
	\setlength{\belowcaptionskip}{-0.cm}
	\centering
	\includegraphics[trim=45 55 0 70, clip,scale=0.3]{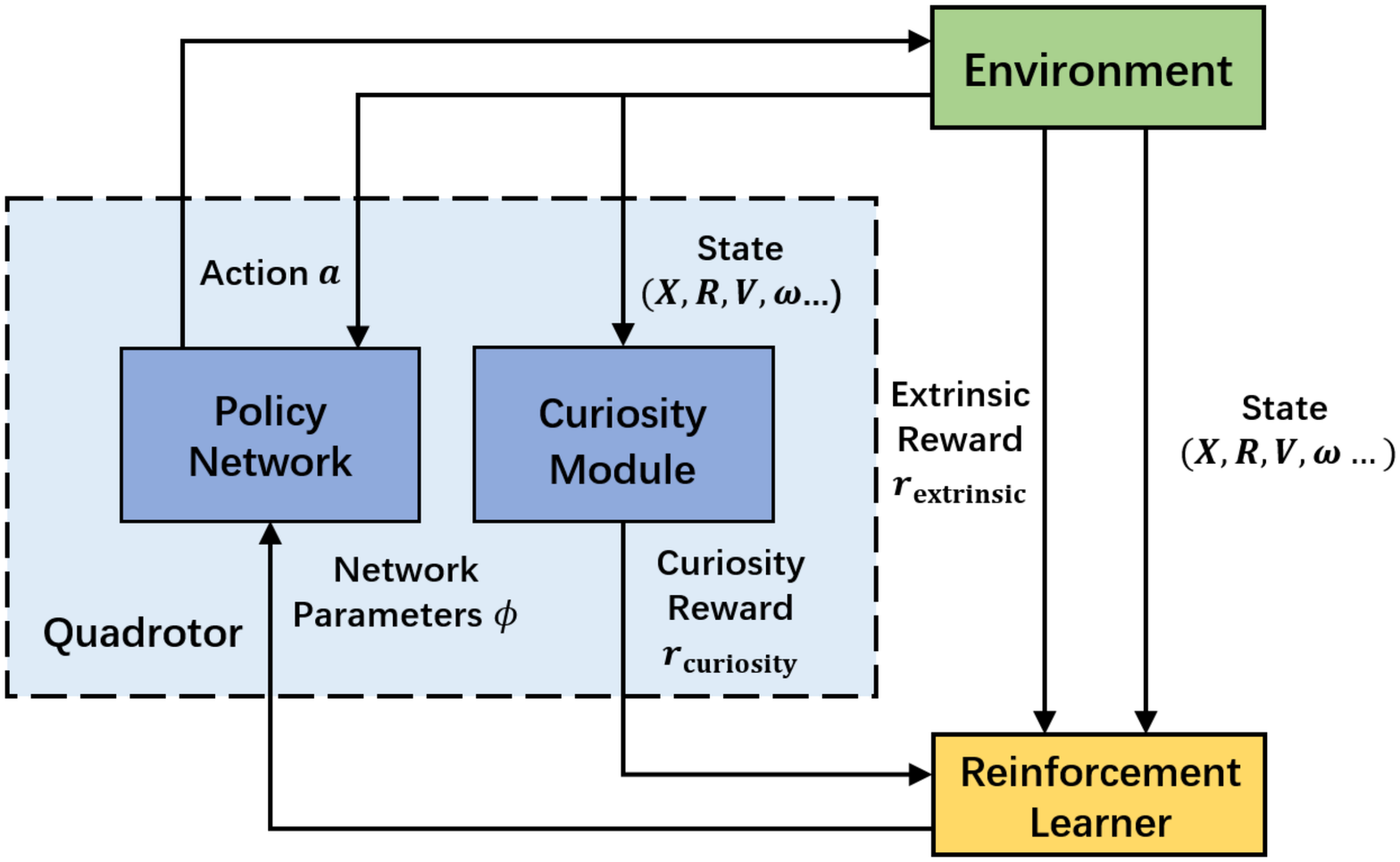}
	\caption{Architecture of our aggressive flight learning process.}
	\label{architecture}
	\vspace{-10pt}
\end{figure}
	
	{
	Though reinforcement learning methods provide solutions to navigation without trajectory planning, few attempts address aggressive flight problems with reinforcement learning. 
	The reason is that reinforcement learning methods still face some challenges when conducting aggressive flight missions. 
	First, the quadrotor system in the aggressive flight missions is underactuated, which brings challenges to the control of the system.
	Thus, positive samples are difficult to obtain by naive stochastic sampling, and the sparse reward issue  becomes more severe~\cite{RLIntro}, which make the exploration of the control policy  difficult and lead  to a severely inefficient training procedure for reinforcement learning. 
	Second, reinforcement learning methods commonly suffer from the transferability issue between simulations and real-world experiments~\cite{Generalization,DomainAdaptation}. A well-performed algorithm in simulations frequently fails in real-world experiments. As these challenges remain unresolved, few works take advantage of the reinforcement learning techniques to address aggressive flight missions.}

%




	{In this work, we explore how to conduct aggressive flight missions, including flying through narrow windows and slalom path scenes, through reinforcement learning using the optimized policy directly instead of tracking the predefined trajectory. The architecture of our algorithm is shown in Figure \ref{architecture}.  
	To obtain more positive samples for training, a curiosity-driven reinforcement learning method is proposed.  
	We use  a curiosity module based on the similarity of the states (i.e., the position and velocity of the quadrotor) for aggressive flight missions to improve the efficiency of policy exploration, overcome the sparse reward issue, and accelerate convergence. 
	To enhance  the adaptability of our method, we set  obstacles with different positions and attitudes  during training, which helps the quadrotor adapt to environments with different  obstacles.
	To ensure that our algorithm can transfer from  simulations to real-world experiments, we add the branch structure exploration (BSE) strategy (also called vine)~\cite{TPRO} to our reinforcement learning algorithm. Through the BSE strategy, we can enrich the diversity of the training samples, which makes the simulation process more similar to the real experiments and enhances the transferability of our method. 	
	In our experiment, we assume that the states of the quadrotor and the obstacles in environment can be detected, and then the quadrotor can make decision according to these states. 	
	The experiments contain two phases: the training phase conducted in simulation and the task execution phase executed in simulation or real-world. 
	In training phase, the quadrotor conducts flight missions in a variety of scenarios with diverse obstacles, which positions and attitudes are different, to learn navigation strategies with our proposed method.
	In task execution phase, the quadrotor executes aggressive flight tasks with the policy trained during the training phase directly, even though the position and attitude of obstacles may be different from those in training phase. 
	Though our method requires an additional training procedure  when compared to traditional methods~\cite{KumarAggressive,ActVision} before conducting aggressive missions, the training procedure is conducted for just one time to enable the system the ability of navigation policy generation. Once the network is well-trained, the quadrotor can execute different aggressive flight missions without extra training.	
	We test our method in both simulations and real scenarios$\footnote{The video of real-world experiments can be found at https://youtu.be/s76iqKB3fNw. The codes of our works can be found at https://github.com/Randy-F/reinforcement-aggressive-flight.}$.} The experimental results show that our method can conduct aggressive flight missions with satisfactory performance in both simulations and real-world experiments, and some ablation experiments demonstrate the effectiveness of our proposed method.

	The main contributions of this study are as follows:
	
	\begin{enumerate}[1)]
		\item {We propose a curiosity-driven reinforcement learning algorithm for aggressive flight missions. Specifically, a similarity-based curiosity module is proposed to obtain more positive samples and overcome the sparse reward problem in reinforcement learning, thus accelerating the convergence speed of the training process.} 
		
		\item The BSE strategy~\cite{TPRO} is applied to guarantee the robustness of the policy, so that the policy trained in simulations can be performed in the real-world directly. {Besides, our method shows the adaptability when the obstacles in a specific aggressive flight mission are different in  test and training process.}
		
		
		\item  Real-world experiments are conducted in both slalom path and narrow window scenes use the policy learned from simulation directly, which demonstrates the simulated to the real (sim2real) transferability of our method is satisfactory. The maximum linear and angular velocity reach 5.65 m/s and 252 deg/s, respectively.		
	\end{enumerate}
	
	\section{RELATED WORK}
	
	Traditional quadrotor navigation and aggressive flight methods generate a feasible trajectory at first and the quadrotor tracks the trajectory with a certain controller~\cite{KumarAggressive,LiuTrajPlan}. In contrast, reinforcement learning quadrotor navigation methods obtain the control strategy by optimizing the policy during the training process~\cite{RL-MPC,MultiFlips}. 
	
	\textbf{Traditional aggressive flight.} Previous works have provided solutions to aggressive flight missions~\cite{KumarAggressive,LiuTrajPlan}. These works focus on the challenges in aggressive flight, including the aggressive trajectory planning~\cite{TIE_plan}, robust controller design~\cite{TIE_control}, and high-rate estimation of aggressive states~\cite{ShenVIO}. In~\cite{KumarAggressive}, the visual-inertial odometry is presented for perception, and an aggressive feasible trajectory planning method is introduced to obtain the sequence of aggressive movements. A lightweight micro aerial vehicle is navigated by this method to fly through some challenging scenes, such as narrow windows or slalom path scenes. The path planning technique with aggressive movements is further discussed in~\cite{LiuTrajPlan}, in which the authors propose a motion-primitive-based graph searching strategy to generate aggressive trajectories in cluttered environments. When the feasibility constraint is presented with an episode quadrotor model, the trajectory can be generated more precisely. {In~\cite{ActVision}, a single onboard camera and an inertial measurement unit (IMU) are used to control a quadrotor and flight through narrow gaps. With the independent yaw-angle planning, the quadrotor is able to continuously face the target while tracking the trajectory.}
	All the methods mentioned above  require a path-planning procedure to generate a desired trajectory, and trajectory planning is very time-consuming, especially for aggressive flight missions. 
	
	Some recent methods~\cite{FastPlan, EventCamera,ryll2019efficient} focus on the faster quadrotor navigation and show significant improvements. In~\cite{FastPlan}, a fast and safe trajectory planner is proposed. The planner can generate trajectories in several seconds and gets about 50$\%$ improvement when compared with other state-of-the-art planning methods~\cite{tordesillas2019real,oleynikova2018safe}. In~\cite{EventCamera}, the event-based cameras are used to navigate quadrotors in complex dynamic environments. The method shows the possibility to accomplish the dynamic obstacle avoidance in few milliseconds and the quadrotors are able to avoid multiple obstacles at relative speeds up to 10m/s. Though~\cite{EventCamera} accomplishes dynamic obstacle avoidance quickly, it utilizes an event-based camera as the sensor  and thus is insensitive to static obstacles, and~\cite{FastPlan} still needs more than 10 seconds for trajectory planning.
	A receding horizon planning architecture is proposed for reactive obstacle avoidance in \cite{ryll2019efficient} and the work is able to perform efficient trajectory planning for high linear speed flight, while our method concentrates on aggressive flight, in which the agents fly with both high linear and high angular speed.
	Compared with these works, our method attempts to avoid the trajectory planning procedure via reinforcement learning methods. By utilizing the end-to-end decision making process, our method saves the  time of planning procedure with the real-time navigation policy.
	
	\textbf{Reinforcement-learning  navigation for quadrotors.} 
	{Reinforcement-learning methods have attracted increasing attention in numerous tasks such as flight control~\cite{FlightControl}, collision avoidance~\cite{CollisionAvoidance} and acrobatics maneuvers~\cite{Acrobatics}. 
	Based on the existence of a system modeling process, these methods can be divided into two categories: model-based~\cite{Trials,Low-level,CollisionAvoidance} and model-free reinforcement learning methods~\cite{RolandRL}.
	Model-based methods model the system dynamics first and then evaluate the policy with these models. In~\cite{Low-level}, the system dynamics model is built by neural networks and is used  to predict future states from past state-action pairs, in which  the  low-level control is performed using a model predictive control strategy. In~\cite{CollisionAvoidance}, the quadrotor is able to avoid obstacles with an uncertainty-aware prediction model after a few training iterations. Model-free  methods are also widely used, and they evaluate the performance of a policy without a system modeling process.} In~\cite{RolandRL}, a deterministic policy optimization method is presented to design a learning-based controller, which can stabilize the quadrotor from a high initial speed and large initial rotation angle.

	
	{Although reinforcement learning methods have shown great  performance in navigation, their sparse reward and poor sim2real transferability remain crucial problems.}
	Sparse reward is a challenging issue in reinforcement learning, and it severely slows down the convergence speed in training process. The curiosity method~\cite{CuriosityPred,CuriosityEps} is proposed as an effective solution to the sparse reward problem.  It encourages the agent to explore the state space more efficiently by providing an additional curiosity reward signal, which is based on the difference between the former and current observations of the agent. In~\cite{CuriosityPred}, a neural network is used to predict the incoming state, and the curiosity reward is generated from the state prediction error. In~\cite{CuriosityEps}, the reachability of episodic states is used as a curiosity signal, and this reachability is estimated by a network approximator.	
	{The sim2real transferability is also an important issue in the reinforcement learning. In~\cite{DomainAdaptation}, the domain adaptation technique is used to improve the transferability of  robotic grasping systems. In~\cite{Generalization}, a quadrotor is navigated in an indoor environment with monocular images, and the experiences in the simulations are used to train a generalizable perception module. Our work relieves the sparse reward issue by a curiosity-driven module and improves the sim2real transferability by the BSE strategy.}


	\section{OUR PROPOSED METHOD FOR AGGRESSIVE FLIGHT}
	
	\subsection{Problem Formulation}

	We describe the aggressive flight missions as a Markov decision process (MDP)~\cite{MDP} and solve it as a standard reinforcement learning problem. There are three important elements in an MDP: state, action, and reward. The quadrotor starts from a certain initial state $s_0$ (such as initial position $\boldsymbol {X}_0$ and velocity $\boldsymbol{V}_0$). By choosing the action $a_t$ (such as attitude $\boldsymbol {R}$ and thrust $f$) by the current policy at time $t$, the states of the quadrotor change from $s_t$ to $s_{t+1}$. The probability distribution of transition $P_{s_t, s_{t+1}}^{a_t}$ is determined by the current state $s_t$ and action $a_t$. The reward $r_t$ is generated by analyzing the quality of the performance caused by the former state $s_t$ and action $a_t$. Assuming that the finite-step process ends in step $T$, a state, action and reward sequence $\{s_t,a_t,r_t\,|\;0\leq t\leq T\}$ is presented. This sequence is used to evaluate the state and action value $Q(s,a)$ and the value updates the action policy $\pi(s)$ by the policy gradient strategy.
	
	We choose the position $\boldsymbol{X}$, attitude $\boldsymbol{R}$ and linear velocity $\boldsymbol{V}$ of the quadrotor along with the parameters of obstacles as the state features of aggressive flight. These features guarantee that the agent obtain sufficient information from the quadrotor and environment. We use the rotation matrix to express the attitude to avoid the discontinuous feature in the $Q$-value approximation~\cite{RLIntro}, which commonly exists when the Euler angle or quaternion is used. The policy network generates the attitude-thrust command, and this command signal is adequate for the requirements of control precision in the aggressive flight mission. Moreover, the attitude-thrust command is a general control command for UAVs with the difference in inner parameters (i.e., dynamic model of the motor), so the policy trained in a simulated environment can obtain a robust performance on a real quadrotor.
	
	To provide guidance for the quadrotor, an extrinsic reward is generated by analyzing the performance in the mission. We divide the extrinsic reward into two parts in the aggressive flight missions: the goal reward providing an encouraging reward and obstacle collision reward providing a punishing reward. 
	\begin{figure}[b]
		\setlength{\abovecaptionskip}{0.cm}
		\setlength{\belowcaptionskip}{-0.cm}
		\centering
		\includegraphics[scale=0.3]{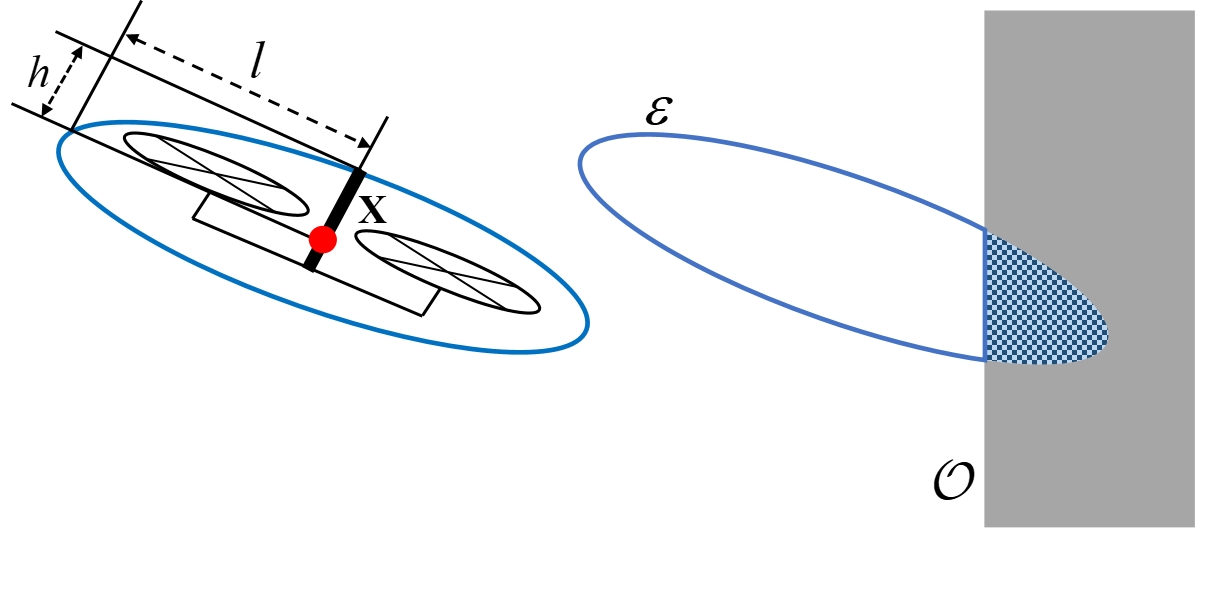}
		\caption{Left: ellipsoid model of quadrotor. Right: illustration of the collision reward generated by the quadrotor model and obstacle.}
		\label{ellipsoid}
	\end{figure}
	The goal reward $r_{\rm goal}$ considers the quality of target movement using the following heuristic function:
	\begin{equation}
	\begin{aligned}
	r_{\rm goal} = \;&\lambda_x |\boldsymbol{X} - \boldsymbol{X}_{\rm goal}| +\lambda_r |\boldsymbol{R} - \boldsymbol{R}_{\rm goal}| \\
	&+ \lambda_v|\boldsymbol{V}| + \lambda_\omega|\boldsymbol{\omega}|,
	\end{aligned}
	\end{equation}
	where $\boldsymbol{X}$, $\boldsymbol{R}$, $\boldsymbol{V}$, and $\boldsymbol{\omega}$ denote the state vectors of position, attitude, linear velocity and angular velocity, $\lambda_x$, $\lambda_r$, $\lambda_v$, and $\lambda_\omega$ denote the weights of the corresponding terms, the subscript ``goal" denotes the reference state of the goal, and the symbol $|\cdot|$ denotes the Euclidean norm. 
	
	The obstacle collision reward is generated by the geometric relationship between the quadrotor and obstacles. We model the quadrotor as an ellipsoid, and this ellipsoid model allows the quadrotor to pass obstacles such as a narrow window with an aggressive attitude~\cite{LiuTrajPlan}. With the known position $\boldsymbol{X}$, attitude $\boldsymbol{R}$, ellipsoid radius $l$, and height $h$, the ellipsoid $\epsilon$ is considered as a point set: 
	
	\begin{equation}
	\epsilon(\boldsymbol{X}, \boldsymbol{R}) := \{\boldsymbol{p} = \boldsymbol{R}\boldsymbol{\Sigma}\boldsymbol{R^T}\boldsymbol{d} + \boldsymbol{X}\, | \; |\boldsymbol{d}|\leq 1\},
	\end{equation}
	where 
	\begin{equation}
	\boldsymbol{\Sigma} = \begin{bmatrix} l & 0 & 0 \\ 0 & l & 0 \\ 0 & 0 & h \end{bmatrix}.
	\end{equation}
	The diagonal matrix $\boldsymbol{\Sigma}$ denotes the quadrotor body configuration, and the vector $d$ assists the description of the point $p$ that meets the conditions. The ellipsoid model is shown in Figure. \ref{ellipsoid}. The obstacle collision reward is: 
	\begin{equation}
	r_{\rm obstacle} = \lambda_{\rm obstacle} \frac {{\rm card}(\epsilon\cap\mathcal{O})}  {{\rm card}(\epsilon)}\,.
	\end{equation}
	This reward is generated by calculating the proportion of the intersection area between ellipsoid $\epsilon$ and obstacle $\mathcal{O}$. $\lambda_{\rm obstacle}$ denotes the weight of the obstacle collision reward, the symbol ${\rm card}(\cdot)$ denotes the number of elements and the set $\epsilon$ and $\mathcal{O}$ are sampled using a point cloud before calculation. The total extrinsic reward is represented as:
	\begin{equation}
	r_{\rm extrinsic} = r_{\rm goal} + r_{\rm obstacle}.
	\label{equation:extrinsic}
	\end{equation}
	

	
	\subsection{Similarity-based Curiosity Module}
	
\begin{figure}[t]
		\setlength{\abovecaptionskip}{0.cm}
		\setlength{\belowcaptionskip}{-0.cm}
		\centering
		\subfigure[]{\centering
			\includegraphics[trim=15 5 0 10, clip, scale=0.65]{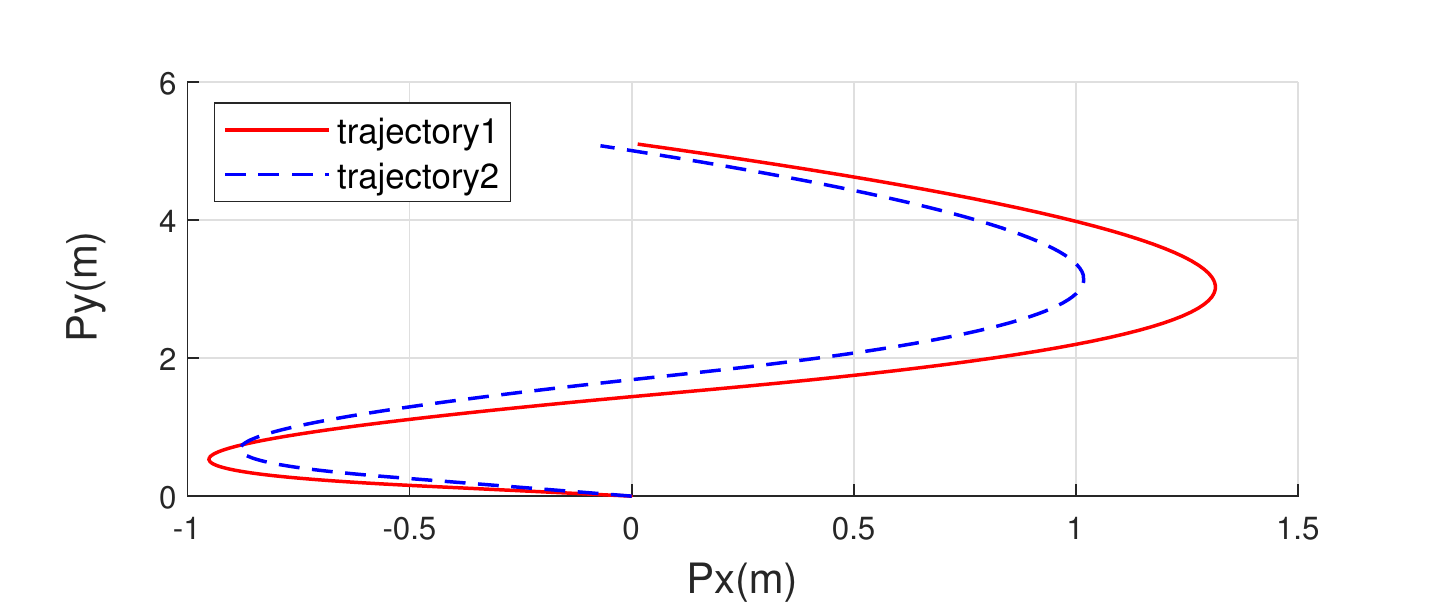}} \quad 
		\subfigure[]{\centering
			\includegraphics[trim=8 0 0 0, clip, scale=0.28]{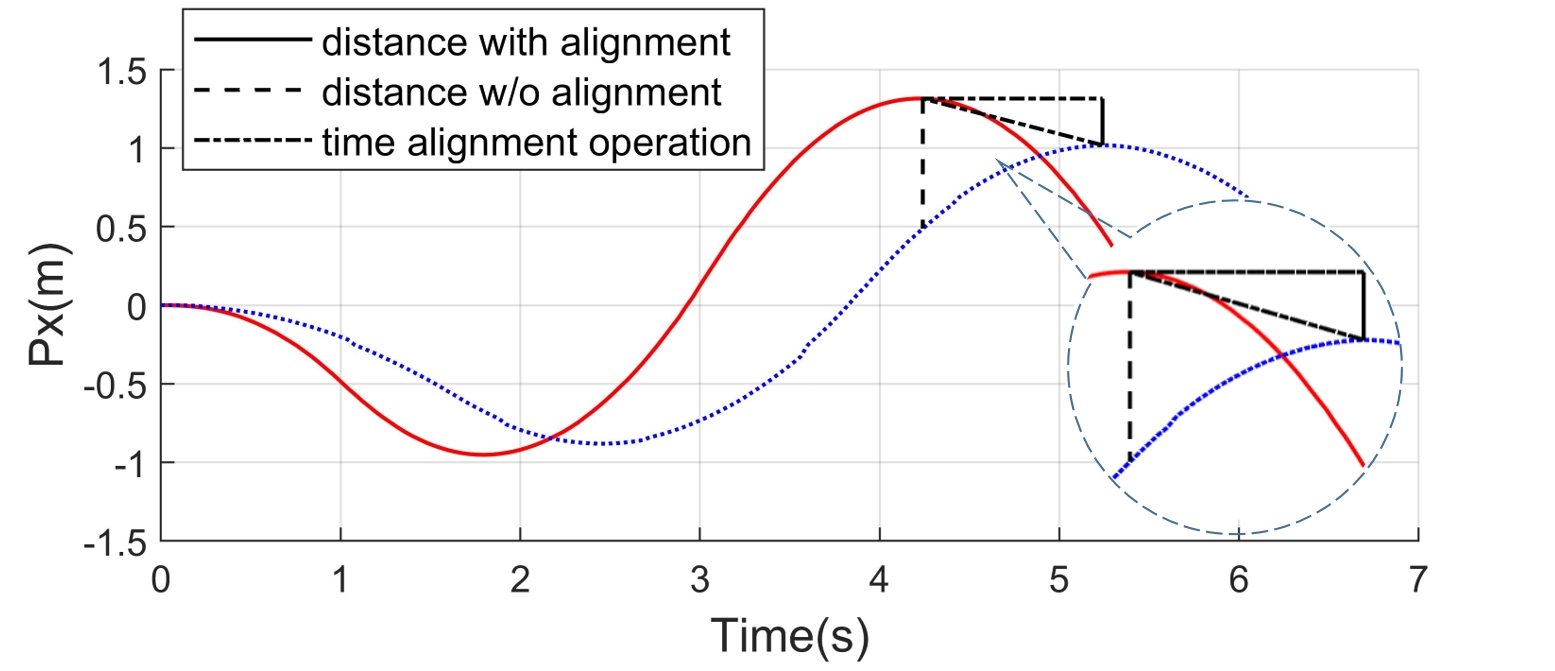}}  
		\caption{Two different trajectory samples: (a) curves of position, (b) curves of time and position\textsf{\textsf{}}. It shows that the similarity measurement without a time alignment (b) is misleading.}
		\label{dtw_pic}
		\vspace{-5pt}
	\end{figure}

	We present a curiosity module to address the sparse reward problem and improve sampling efficiency. In an aggressive flight mission, the quadrotor is required to perform movements in extreme conditions. Therefore, a successful reward is very difficult to obtain by simply using a random sampling strategy, which makes the sparse reward problem appear. To address this problem and encourage the agent to experience novel states, we propose a similarity-based curiosity module for the aggressive flight mission. Unlike using reachability checking or state prediction error, our curiosity module judges the similarity of the time-state curves between the current and former episodes. This method describes the similarity relationship between different episodes more precisely and makes full use of the information in the entire training process. Considering a low similarity as a high curiosity reward, the similarity of the states between different episodes provides a guideline to lead the quadrotor to explore new states. However, a simple comparison of similarities without a proper time alignment is misleading. For example, in Figure  \ref{dtw_pic}, two curves of the states during different training episodes are shown to demonstrate the effect of the time alignment. Note that in Figure  \ref{dtw_pic}(a), these two episodes have quite similar positional trajectories; however, the measurements is with highly dissimilar when we directly compare them in the time-state curves in Figure  \ref{dtw_pic}(b). 
	


	Thus, we perform time alignment operation before the comparison to make  the similarity measurement between these curves  more accurate.	
	Assume that the quadrotor performs an aggressive flight several times and $S=\{s_i \,|\; i=1,2,...,n \}$, $S'=\{s'_j \,|\; j=1,2,...,m \}$ are two of the state sequences in these episodes. 
	To perform a proper time alignment, we use the dynamic time warping method~\cite{DTW} for the curve similarity measurement with matrix $\textbf{A}_{n\times m}$. Its element $a_{(i,j)}$ is the  distances between each state ${s_i}$ and ${s'_j}$ in episode $S$ and $S'$. 
	Then, the minimized distance with time alignment can be obtained by finding a set of consecutive elements  in the matrix and minimizing the sum of the all the elements  in the set. Specifically, in matrix $\textbf{A}_{n\times m}$, a set of elements  starting from the top left corner and ending in the bottom right corner denotes a possible choice for time alignment, and the element $a_{(i,j)}$ in the chosen set denotes that the states $({s_m}$, ${s'_j})$ are aligned.	
	We can obtain the warping set by iteratively calculating the aligned distance:
	
	\begin{equation}
	D(s_i, s'_j)=d(s_i, s'_j)+ \min(\mathcal{D}'_{i,j}), 
	\end{equation}
	where 
	\begin{equation}
	\mathcal{D}'_{i,j} = \{D(s_{i-1}, s'_{j-1}), D(s_{i-1}, s'_j), D(s_i, s'_{j-1})\},
	\end{equation}
	and $D(s_i, s'_j)$ denotes the accumulated distance from state $(s_1,s'_1)$ to $(s_i, s'_j)$ along the warping set. With this expression, we can get the distance between $S$ and $S'$ using dynamic time warping (DTW):
	\begin{equation}
	D_{\rm dtw}(S,S') = D(s_n,s'_m)\,.
	\end{equation}
	
	
	The curiosity reward is generated based on this aligned similarity measurement. It is calculated by the minimum state distance when the current episode is compared with each of the former episodes. The final curiosity reward is
	\begin{equation}
	r_{\rm curiosity} = 1-{\rm exp}(-\min\limits_{i} \{ \sum\limits_{n=1}^N D_{\rm dtw}(S_n, {S'}^i_n)\}),
	\label{equation:curiosity}
	\end{equation}
	where $S_n$ denotes the $n$th state in the current episode,  ${S'}^i_n$ denotes the $n$th state in the $i$th former episode, and we use position, attitude and velocity for similarity measurement.
	We use the negative exponential function to restrict the value of the curiosity reward to a reasonable range. The similarity-based curiosity module encourages state space exploration and reduces the time cost of the training time with high-quality samples. We generate the total reward function  
	\begin{equation}
	r = r_{\rm extrinsic} + \lambda_{c}r_{\rm curiosity},
	\end{equation}
	where $\lambda_{c}$ is the weight of $r_{\rm curiosity}$.

	\subsection{Exploration Strategy}
	
	We use the BSE strategy to improve the robustness of the policy~\cite{TPRO}. During the process of convergence in the training procedure, the agent tends to choose the fixed action sequence, and the trajectories of the quadrotor in each episode will be similar to each other. In simulations, this type of strategy is able to work and ensure a satisfactory result. However, when it comes to real-world environments, the performance of this kind of strategy is far from satisfactory. Because there exist different kinds of ubiquitous and unavoidable disturbances (e.g., actuator bias or uncertain airflow) exist in reality and these disturbances are difficult to consider in simulations, the quadrotor will experience new states and act inappropriately due to these disturbances when real-world experiments are conducted. 
	
	\begin{figure}[t]
		\setlength{\abovecaptionskip}{0.cm}
		\setlength{\belowcaptionskip}{-0.cm}
		\centering
		\includegraphics[scale=0.28]{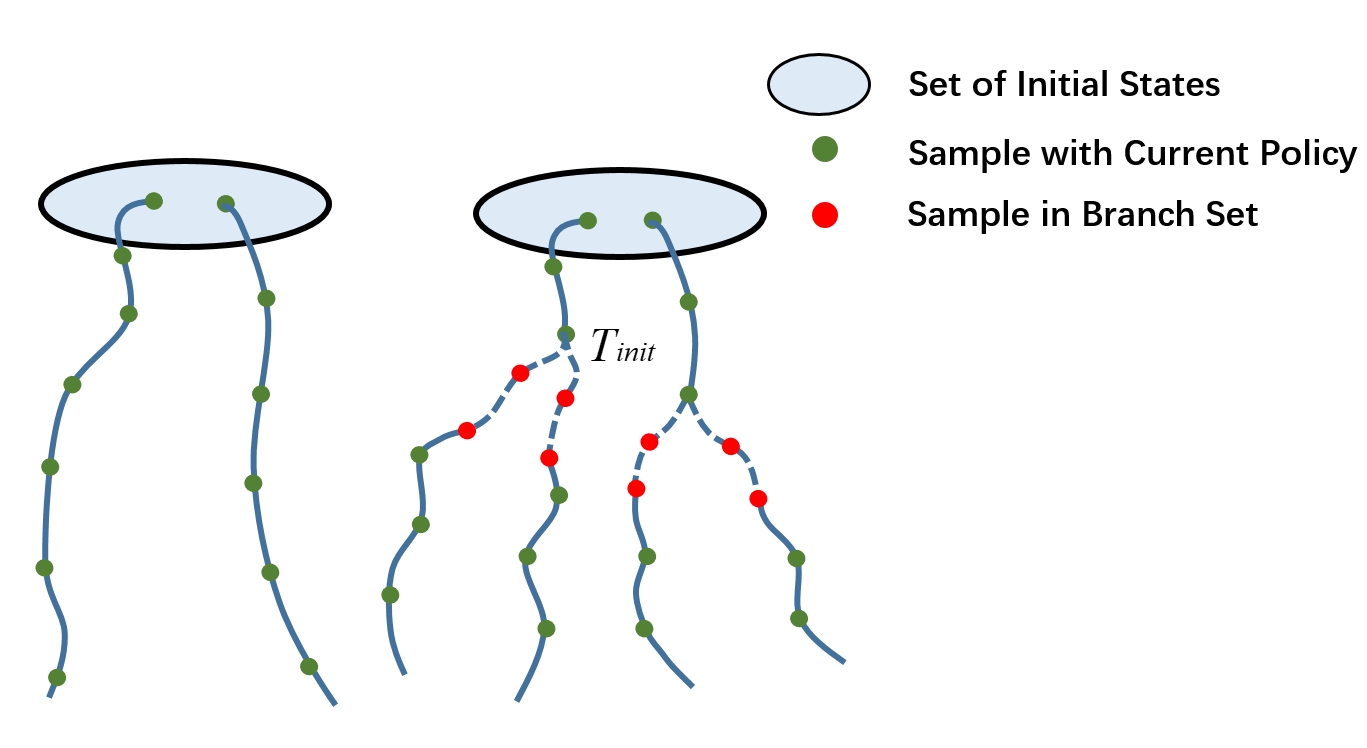}
		\caption{Single path exploration method (left) and BSE method (right). The BSE obtains a broader sampling area.}
		\label{exploration}
		\vspace{-5pt}
	\end{figure}
	
	Therefore, to extend the exploration area and improve the quality of the action in a wider state space, we perform an exploration strategy in the branch structure. The BSE is illustrated in Figure \ref{exploration}. We start the experiments with the current policy $\pi_\phi$, obtaining the initial samples $s_0, s_1, ..., s_{_{T_{init}}}$. Then, a branch set of the reachable states are generated following the initial samples. For the states in this branch set, we perform an action with additional random noise to generate a branch trajectory. The following trajectory sampled with the current policy starts from the end of the branch trajectory. Compared with the simple single path exploration, the BSE method guarantees a broader distribution of sampling area. 
	{Thus, the robustness of the control policy is improved as well.}
	
	\subsection{Learning-based Policy Generation}
	Our learning-based aggressive flight architecture is shown in Figure  \ref{architecture}.  We introduce a curiosity module for the training procedure to generate an intrinsic curiosity reward. The agent performs actions in an aggressive flight environment, and the actions are selected by the policy network using the state information. The reinforcement learner also receives the state information, with the extrinsic reward in Eq. (\ref{equation:extrinsic}) and the intrinsic curiosity reward from our similarity-based curiosity module in Eq. (\ref{equation:curiosity}). During the learning process, the reinforcement learner evaluates the policy using the former state, action, and reward sequence and updates the policy network with new network weights. 
	
\begin{figure}[t]
		\centering
		\includegraphics[scale=0.21]{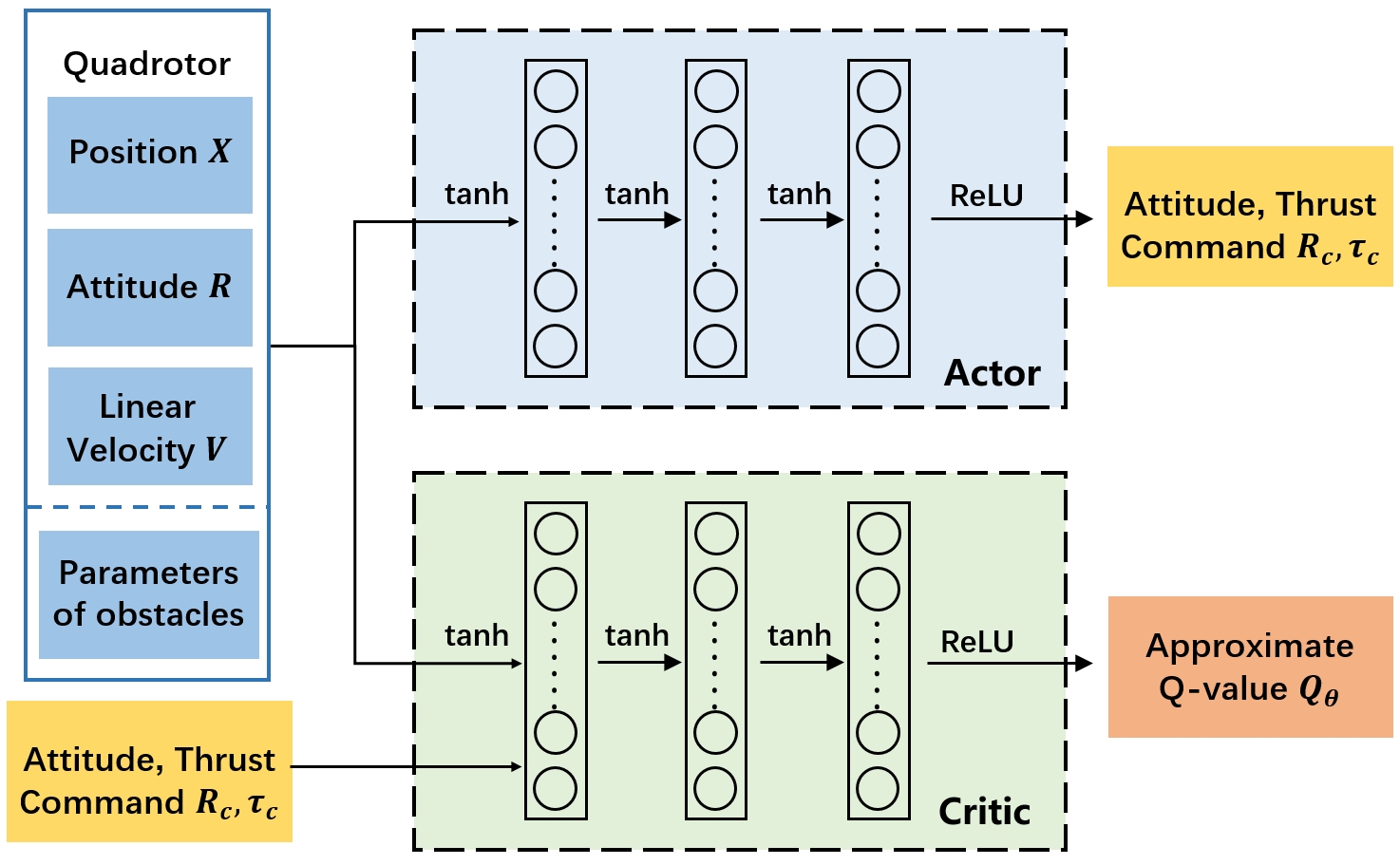}
		\caption{Illustration of the network structure, both the actor and critic network have a similar structure with different input and output.}
		\label{network}
		\vspace{-5pt}
	\end{figure}

	In this learning-based aggressive flight architecture, three networks are included to optimize the policy: the actor network (policy network), the critic network, and the target network. The actor network generates actions through state information; the critic network evaluates the value of states and actions; and the target network approximates the $Q$-value $Q(s,a)$ as a supervised signal of the critic network. The structures of actor and critic networks are shown in Figure  \ref{network}. Our experiments show that this simple network structure provides satisfactory performance with our learning method. The target network is designed with the same structure as the entire actor-critic network to evaluate the target value~\cite{HumanLevelCtrl}.
	
		\begin{figure*}[t]
		\setlength{\abovecaptionskip}{0.cm}
		\setlength{\belowcaptionskip}{-0.cm}
		\centering
		\subfigure[The narrow window mission.]{\includegraphics[trim=22 0 0 10, clip, scale=0.6]{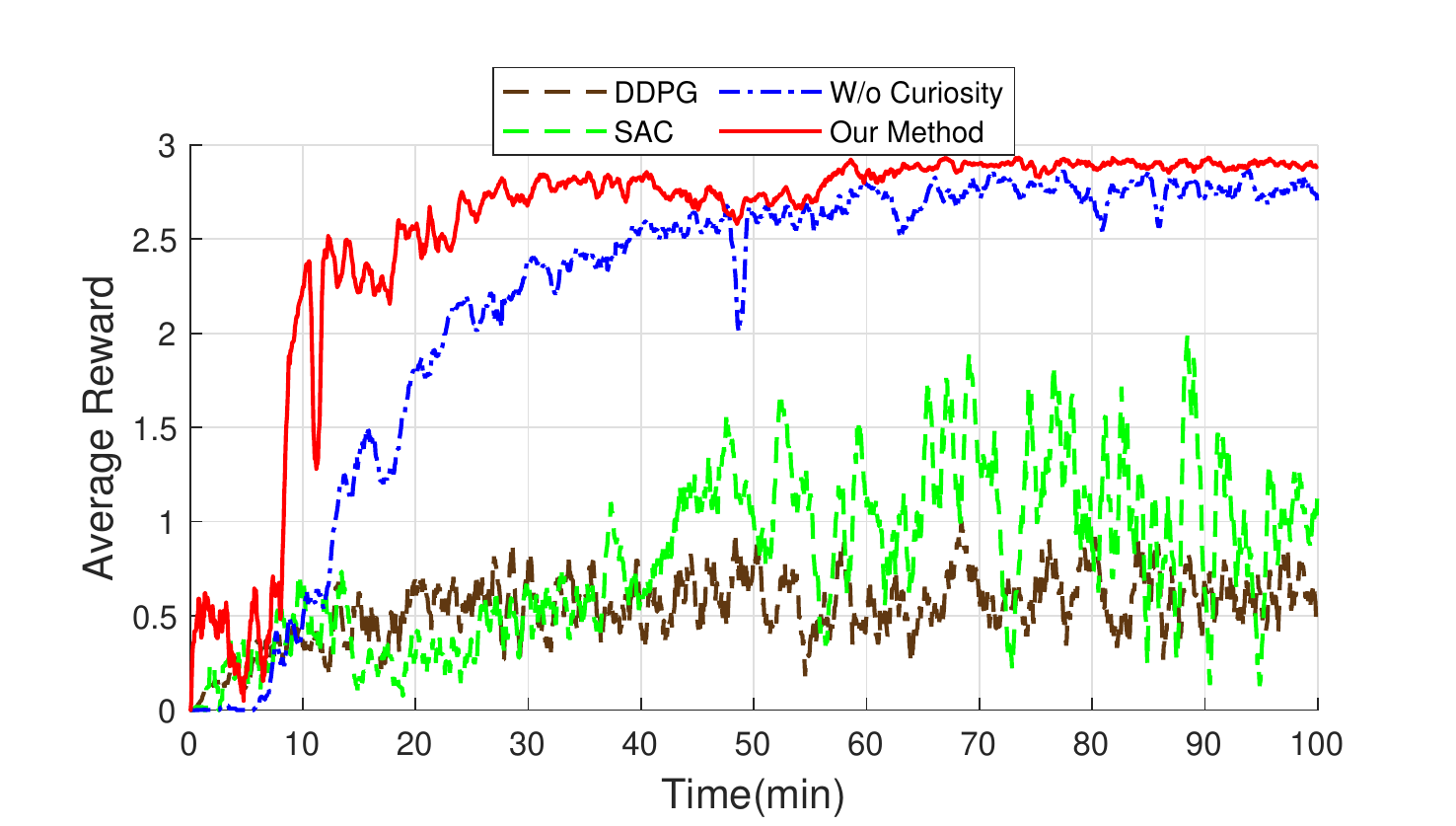}} \quad 
		\subfigure[The slalom path mission.]{\includegraphics[trim=24 0 0 8, clip, scale=0.6]{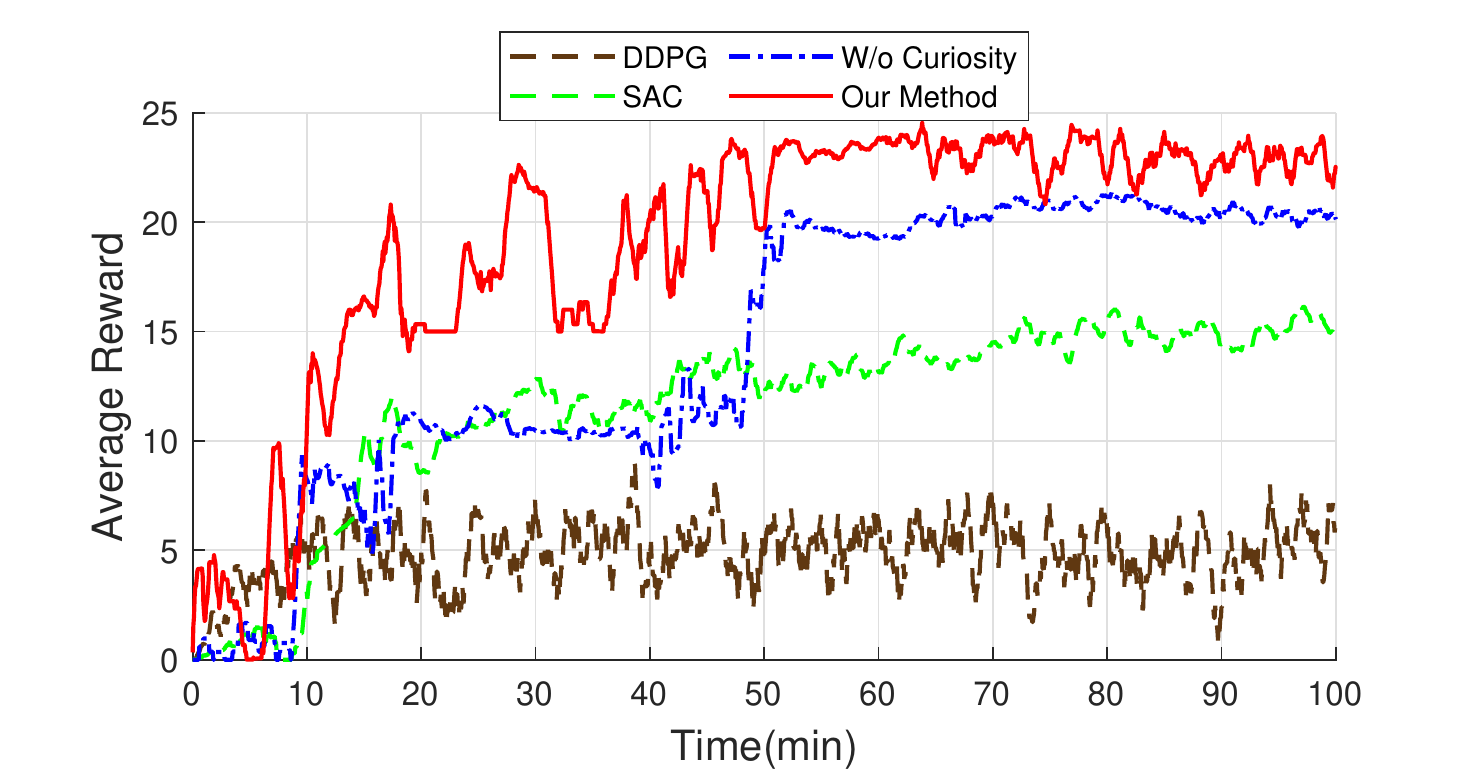}}  
		\caption{Learning curves of our method, our method without curiosity module, SAC and DDPG in aggressive flight missions.}
		\label{learning_curve}
		\vspace{-5pt}
	\end{figure*}
	
	
	\begin{algorithm}[b]
		\caption{Curiosity-Driven Deterministic Policy Gradient}
		\label{alg:A}
		\begin{algorithmic}
			\STATE Initialize critic-actor networks and target networks
			\STATE Initialize the replay buffer $\mathcal{B}$ and delayed parameter $d$
			\STATE Observe $s_0$
			\FOR {$t = 0,1,2,...,T$}
			\STATE Sample action $a_t$ using current policy
			\IF {episode is terminated}
			\STATE	Calculate curiosity reward $r_{curiosity}$
			\ENDIF
			\STATE Observe reward $r_t$ and new state $s_{t+1}$
			\STATE Store transition ($s_t$, $a_t$, $r_t$, $s_{t+1}$) in $\mathcal{B}$ 
			\STATE Sample $N$ transitions from $\mathcal{B}$
			\STATE Update critics using gradient descent step
			\IF {$t\; \rm{mod}\; d == 0$}
			\STATE	Update actor network according to Eq. \ref{actor_eq}
			\STATE	Update target network according to Eq. \ref{target_eq}
			\ENDIF
			\ENDFOR
		\end{algorithmic}
	\end{algorithm}
	
	Our learning process is based on the deep deterministic policy gradient (DDPG)~\cite{DDPG}. To address the over estimation problem in $Q$-value approximation~\cite{over-estimation}, we adopt the twin delayed deep deterministic policy gradient (TD3)~\cite{TD3} method. A pair of critic networks are introduced with corresponding target networks, and a delayed policy update strategy is adopt. With these two target networks individually providing the estimation of the target value $G_t(s, a)$, we update the target value by using the smaller estimation:
	
	\begin{equation}
	G_t(s_t, a_t) = v_{t} + \gamma*\min\limits_{i=1,2}Q_{\theta'_i}(s_{t+1}, \pi_{\phi'}(s_{t+1})),
	\end{equation}
	where $G_t(s_t, a_t)$ is the target estimation of state $s_t$ and action $a_t$, and $v_{t}$ is the preliminary approximation of the value in the $t^{th}$ step using Monte-Carlo samples~\cite{RLIntro}. $\gamma$ is the discount factor~\cite{RLIntro} in the MDP and $\theta'_1$, $\theta'_2$, and $\phi'$ are the weights of the target actor-critic networks.
	The policy network is updated using the DDPG~\cite{DDPG}:
	\begin{equation}\\\label{actor_eq}
	\nabla_\phi L(\phi) = \frac 1N \sum\limits_{k=0}^N  \nabla_aQ_{\theta_ 1}(s,a)|_{a=\pi_\phi(s)}\nabla_\phi\pi_\phi(s),
	\end{equation}
	where $\phi$ is the weights of the actor network, $N$ is the mini-batch sample size and $k$ is the sample index. Finally, we update the target network by 
	
	\begin{equation}\label{target_eq}
	\begin{aligned}
	\phi' \leftarrow \rho\phi + (1-\rho)\phi', \\
	\theta'_i \leftarrow \rho\theta_i + (1-\rho)\theta'_i,
	\end{aligned}
	\end{equation}
	with the same frequency as the policy update. To stabilize the learning process, we update the policy and target networks when the critic networks are updated more than once. This delayed policy update strategy guarantees a small estimation bias in the policy update process.
	Our curiosity-driven deterministic policy gradient algorithm is summarized in Algorithm 1.
	

	\section{EXPERIMENTAL RESULTS}
	
	\subsection{Simulation Setup}
	 Unreal Engine 4$\footnote{https://www.unrealengine.com/}$ is used as the physical engine to simulate the dynamic model of quadrotor and training environment. The simulation platform is equipped with a 2.20GHz i7-8750H CPU, NVIDIA GTX 1070 GPU, and 16GB memory space. The training process is implemented in an Airsim$\footnote{https://github.com/Microsoft/AirSim}$ aerial vehicle simulator. We implement our proposed reinforcement learning algorithm in two aggressive flight missions, flying through the slalom path and a narrow window. The slalom path scene contains two columnar obstacles, and the  obstacles are in front of the initial position of the agent. The goal of the task is to pass through both obstacles in opposite horizontal directions. For the narrow window scene, the goal of the task is to pass through the narrow window gap with a certain attitude. We terminate a training episode when the target is accomplished or the quadrotor is outside the flight region. The curiosity reward is generated when an episode is terminated, the extrinsic reward is generated when a collision is detected or the target is reached, and we set $\lambda_{c}$ = 4.
	
	
	\begin{table}[t]
	\vspace{-1pt}
	\captionsetup{justification=centering}
	\caption{Statistics in narrow window scene with different setup of noises.}
	\label{table_1}
	\begin{center}
		\renewcommand{\multirowsetup}{\centering}
		\begin{tabular}{m{40pt}<{\centering}m{44pt}<{\centering}m{34pt}<{\centering}m{28pt}<{\centering}m{31pt}<{\centering}}
			\hline\hline
			STD noise & method &Position error (m)& Average reward& Successful rate\\
			\hline
			\specialrule{0em}{0pt}{1pt}
			\multirow{2}{*}{$0.0^{\circ}$} & \textbf{Our method} &\textbf{0.0116} & \textbf{2.9154}& \textbf{99.2$\%$}\\ \specialrule{0em}{0pt}{1pt}
			\specialrule{0em}{0pt}{1pt}
			& SPE method  &0.0158 & 2.9045& 98.6$\%$\\ \specialrule{0em}{0pt}{1pt}
			\specialrule{0em}{0pt}{1pt}
			\multirow{2}{*}{$1.5^{\circ}$} & \textbf{Our method} &\textbf{0.0138} & \textbf{2.9067}& \textbf{96.8}$\%$\\ \specialrule{0em}{0pt}{1pt}
			\specialrule{0em}{0pt}{1pt}
			& SPE method &0.0219 & 2.8973& 79.5$\%$\\ \specialrule{0em}{0pt}{1pt}
			\specialrule{0em}{0pt}{1pt}
			\multirow{2}{*}{$3.0^{\circ}$}  & \textbf{Our method} &\textbf{0.0164} & \textbf{2.8973}& \textbf{91.2}$\%$\\ \specialrule{0em}{0pt}{1pt}
			\specialrule{0em}{0pt}{1pt}
			& SPE method &0.0424 & 2.7854& 50.6$\%$\\ \specialrule{0em}{0pt}{1pt}
			\hline\hline
		\end{tabular}
	\end{center}
\end{table}

\begin{table}[t]
	\setlength{\abovecaptionskip}{0pt}
	\setlength{\belowcaptionskip}{10pt}
	\captionsetup{justification=centering}
	\caption{Statistics in slalom path scene with different setup of noises.}
	\label{table_2}
	\begin{center}
		\renewcommand{\multirowsetup}{\centering}
		\begin{tabular}{m{40pt}<{\centering}m{44pt}<{\centering}m{34pt}<{\centering}m{28pt}<{\centering}m{31pt}<{\centering}}
			\hline\hline \\[-3mm]
			STD noise & method &Position error (m)& Average reward& Successful rate\\
			\hline
			\specialrule{0em}{0pt}{1pt}
			\multirow{2}{*}{$0.0^{\circ}$} & \textbf{Our method} &\textbf{0.0149} & \textbf{24.179}& \textbf{99.1$\%$}\\ \specialrule{0em}{0pt}{1pt}
			\specialrule{0em}{0pt}{1pt}
			& SPE method  &0.0163 & 24.096& 98.8$\%$\\ \specialrule{0em}{0pt}{1pt}
			\specialrule{0em}{0pt}{1pt}
			\multirow{2}{*}{$1.5^{\circ}$} & \textbf{Our method} &\textbf{0.0158} & \textbf{23.98}5& \textbf{96.5$\%$}\\ \specialrule{0em}{0pt}{1pt}
			\specialrule{0em}{0pt}{1pt}
			& SPE method &0.0192 & 23.885& 86.3$\%$\\ \specialrule{0em}{0pt}{1pt}
			\specialrule{0em}{0pt}{1pt}
			\multirow{2}{*}{$3.0^{\circ}$}  & \textbf{Our method} &\textbf{0.0185} & \textbf{23.829}& \textbf{93.6$\%$}\\ \specialrule{0em}{0pt}{1pt}
			\specialrule{0em}{0pt}{1pt}
			& SPE method &0.0254 & 23.429& 61.2$\%$\\ \specialrule{0em}{0pt}{1pt}
			\hline\hline
		\end{tabular}
	\end{center}
\end{table}
	
	We conduct the training procedure in both slalom path and narrow window scenes, and we compare our methods with DDPG~\cite{DDPG} and the Soft Actor-Critic (SAC)~\cite{SAC}. To prove the effectiveness of our curiosity module on the convergence speed, ablation experiments are conducted as well. Following previous works~\cite{DDPG,SAC}, we adopted the average reward as the evaluation criterion for the performance of the reinforcement learning algorithms. 
	From the average reward learning curves shown in Figure  \ref{learning_curve}, it is obvious that our method achieves the best performance when compared to DDPG~\cite{DDPG} and SAC~\cite{SAC}. The method without the curiosity module takes more time for training and obtains lower rewards, showing that our curiosity module improves the exploration efficiency and performance.	
	In particular, in the narrow window mission, our method obtains a satisfactory  reward in about 20 minutes of training and the average reward learning curve converges in about 30 minutes.  In the slalom path mission, our method reaches convergence in 45 minutes, while other methods are not able to achieve the same performance with even in 100 minutes. Facing multiple sequential targets in a slalom path scene, our method performs much better than the others in terms of exploration efficiency.

	{Since  reinforcement learning methods are  poor in sim2real transferability, we use the BSE strategy~\cite{TPRO} to improve the robustness of our system and to improve the sim2real transferability. 
	The disturbances in real-world experiments could come from the voltage instability of batteries, nonstandard dynamics of rotors, uneven mass distribution, etc., which lead to the actions (attitude commands in our experiments) in real experiments deviating from the simulated cases. Therefore, we add action noises to simulate the uncertainty in real-world experiments. The action noises are denoted by the attitude command error and are applied to each attitude axis independently. In the experiments, we set the standard deviation (STD) noises at $0.0^{\circ}$, $1.5^{\circ}$, and $3.0^{\circ}$.} 

\begin{figure}[t]
	\setlength{\abovecaptionskip}{0.cm}
	\setlength{\belowcaptionskip}{-0.2cm}
	\centering
	\includegraphics[trim=17 0 0 10, clip,scale=0.6]{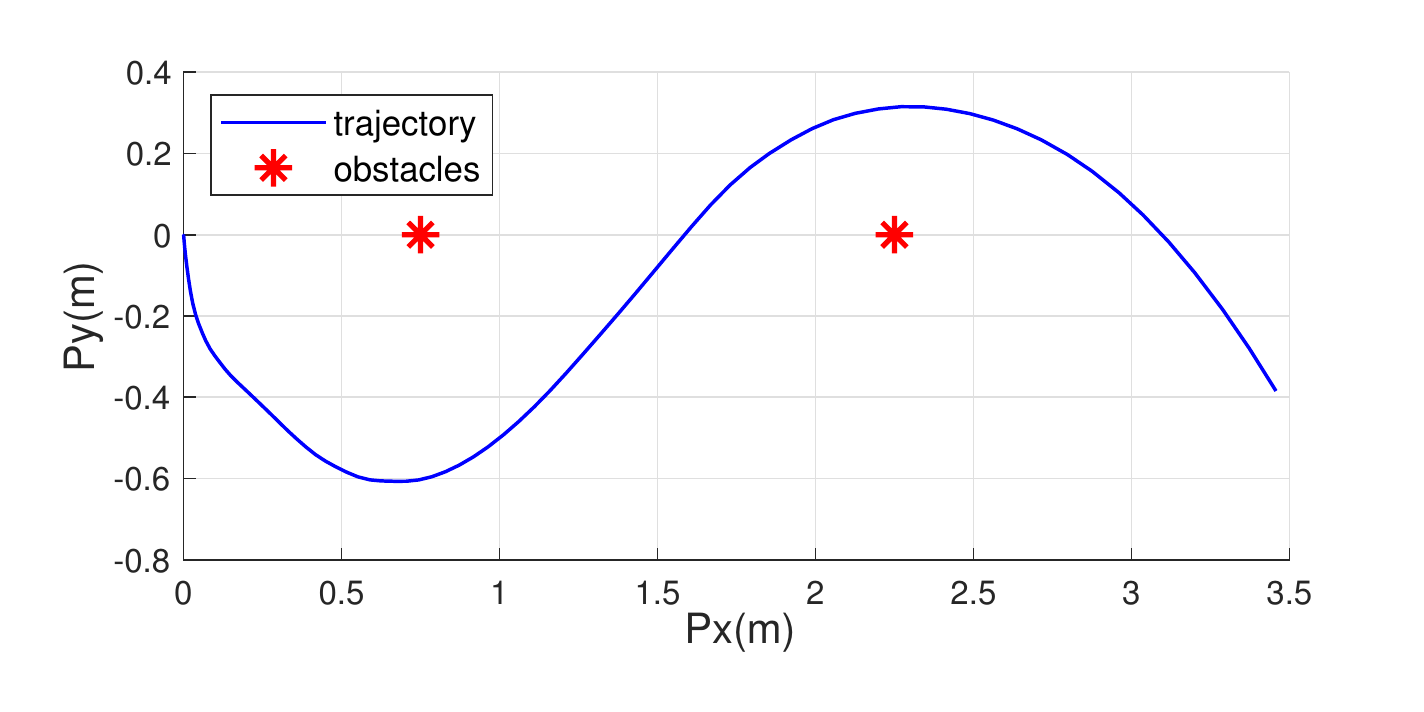}
	\caption{Position trajectory in slalom path scene.}
	\label{slalomtraj_sim}
	\vspace{-5pt}
\end{figure}

\begin{figure}[t]
	\setlength{\abovecaptionskip}{0.cm}
	\setlength{\belowcaptionskip}{-0.25cm}
	\centering
	\subfigure[Position trajectories.]{\includegraphics[trim=20 15 0 0, clip, scale=0.6]{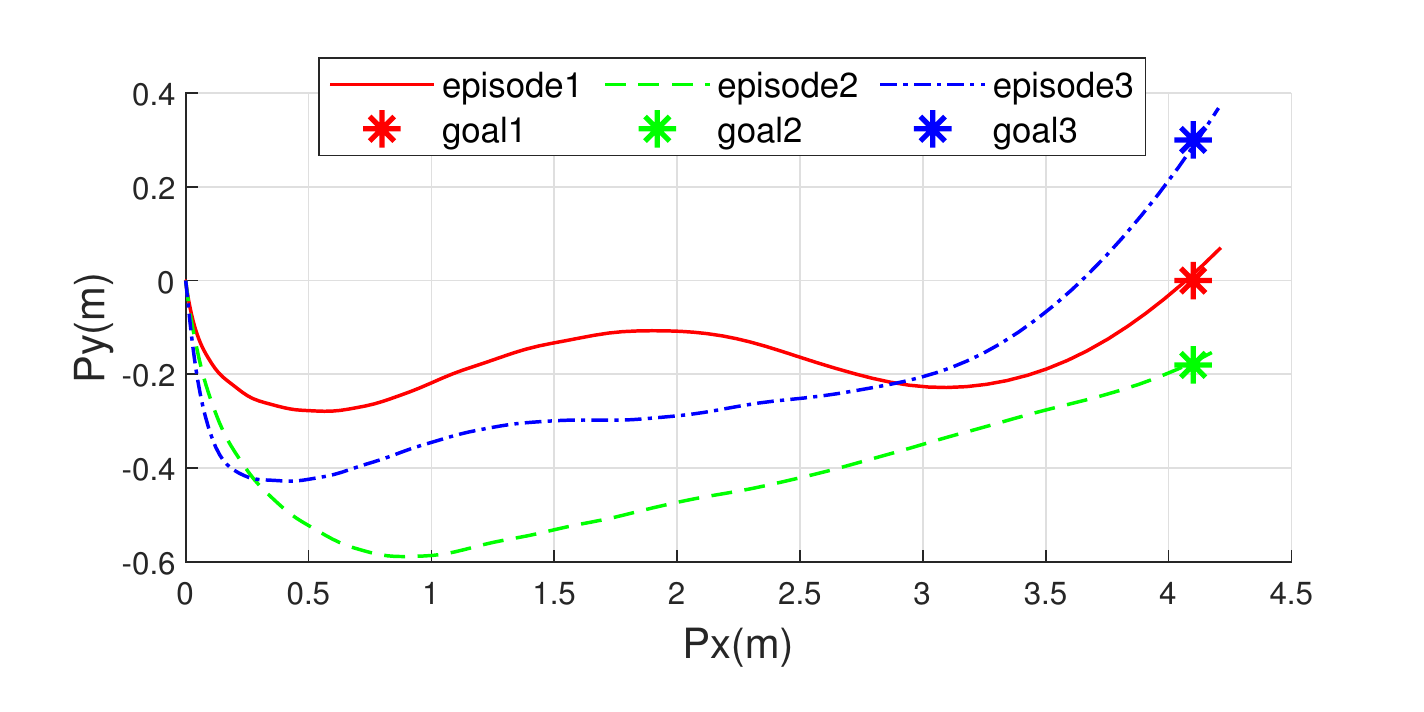}} \quad 
	\subfigure[Rotation trajectories.]{\includegraphics[trim=20 18 0 10, clip, scale=0.6]{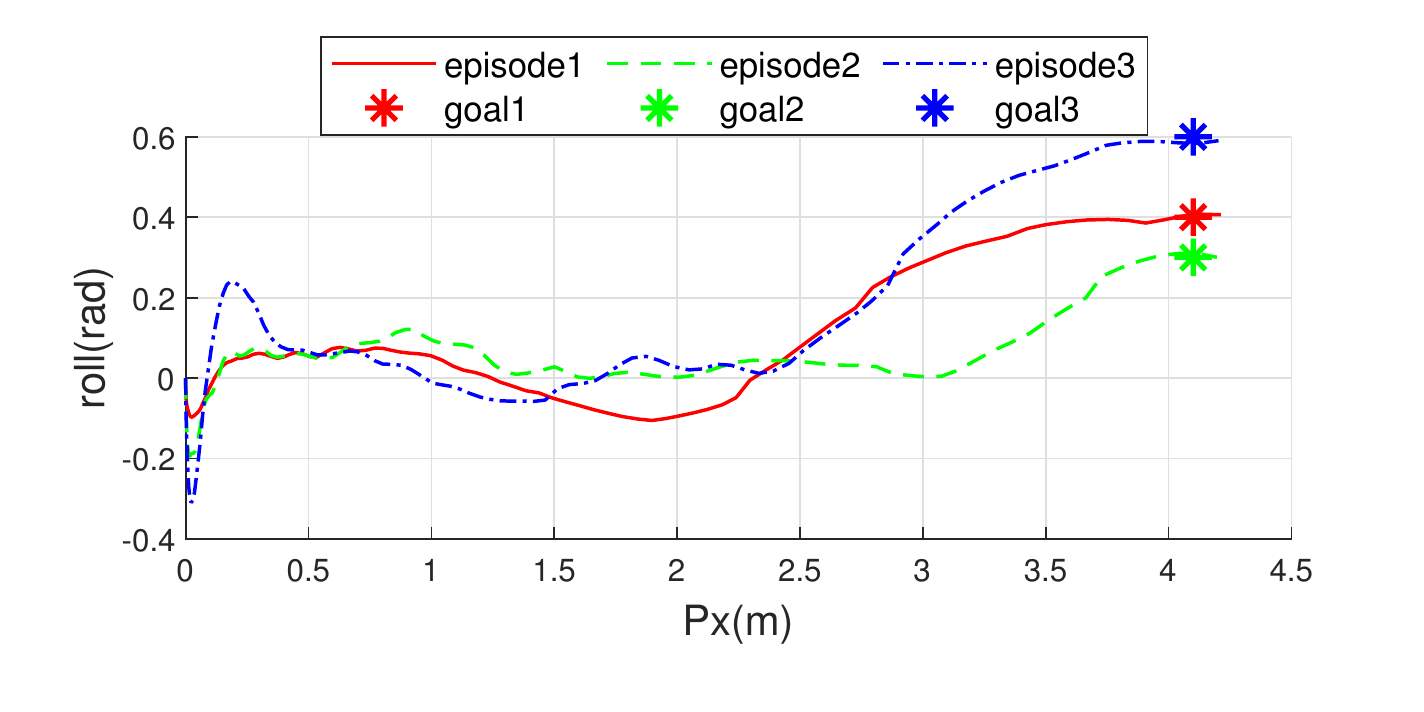}}
	\caption{Position  and rotation  trajectories in narrow window scene.}
	\label{wintraj_sim}
	\vspace{-5pt}
\end{figure}

	The quantitative experimental results are shown in Tables \ref{table_1} and \ref{table_2}. We also conduct an ablation experiment to verify the effectiveness of the BSE strategy by replacing it with the single path exploration method (SPE method)~\cite{TPRO} and test both of the methods with additional action noises.  When the STD noise is 1.5$^{\circ}$, our method maintains the accuracy while the error of the SPE method increases significantly. When the STD of noise is 3$^{\circ}$, the SPE method tends to fail while our method has a much higher success rate owing to its wider sample border. Each experiment was performed 1000 times. The trajectories in the simulation experiments are demonstrated in Figure  \ref{slalomtraj_sim} and Figure  \ref{wintraj_sim}. The trajectory in the slalom path scene is shown in Figure  \ref{slalomtraj_sim}, in which we choose an episode to show the performance of the policy. {The trajectories in different narrow window scenes are plotted in Figure  \ref{wintraj_sim}, in which we perform the experiments on diverse narrow windows using the same pre-trained policy. The narrow window scenes are different in the various rotation angles and the distances to the center of the scenes. Specifically, the angles of these windows are 0.3, 0.4 and 0.6 rad, and the distances are -0.18, 0.0 and 0.3 m, respectively.} The figures demonstrate the generalization of our method to navigate the quadrotor with variable obstacles parameters. 
	

To further verify the superiority of our method, we run a Monte-Carlo simulation for a collection of scenes containing random unstructured obstacles. 
	In the generated scenes, there exists at least one path that is in accord with quadrotor's dynamics constraints for the agent to traverse across.
	We execute navigation in the generated scenes using our method and an example of the experimental result is shown in Figure 9. The result demonstrates that our method not only performs well in classical aggressive flight missions like specific narrow window and slalom path scenarios, but also works well in other unstructured environments. More experiments can be found in Appendix A.

	\begin{figure}[t]
	\vspace{-5pt}
	\setlength{\abovecaptionskip}{0.2cm}
	\setlength{\belowcaptionskip}{-0.cm}
	\centering
	\includegraphics[scale=0.7]{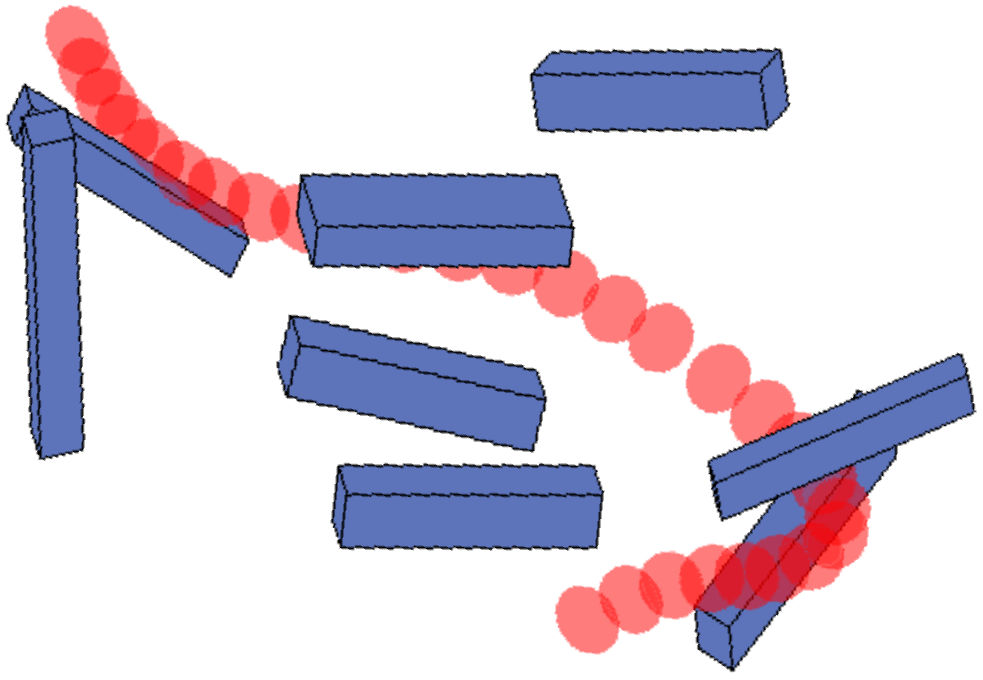}
	\caption{The trajectory in an unstructured environment.}
	\label{unstructered}
	\vspace{-5pt}
\end{figure}

\begin{figure}[t]
	\vspace{-5pt}
	\setlength{\abovecaptionskip}{0.cm}
	\setlength{\belowcaptionskip}{-0.cm}
	\centering
	\includegraphics[scale=0.36]{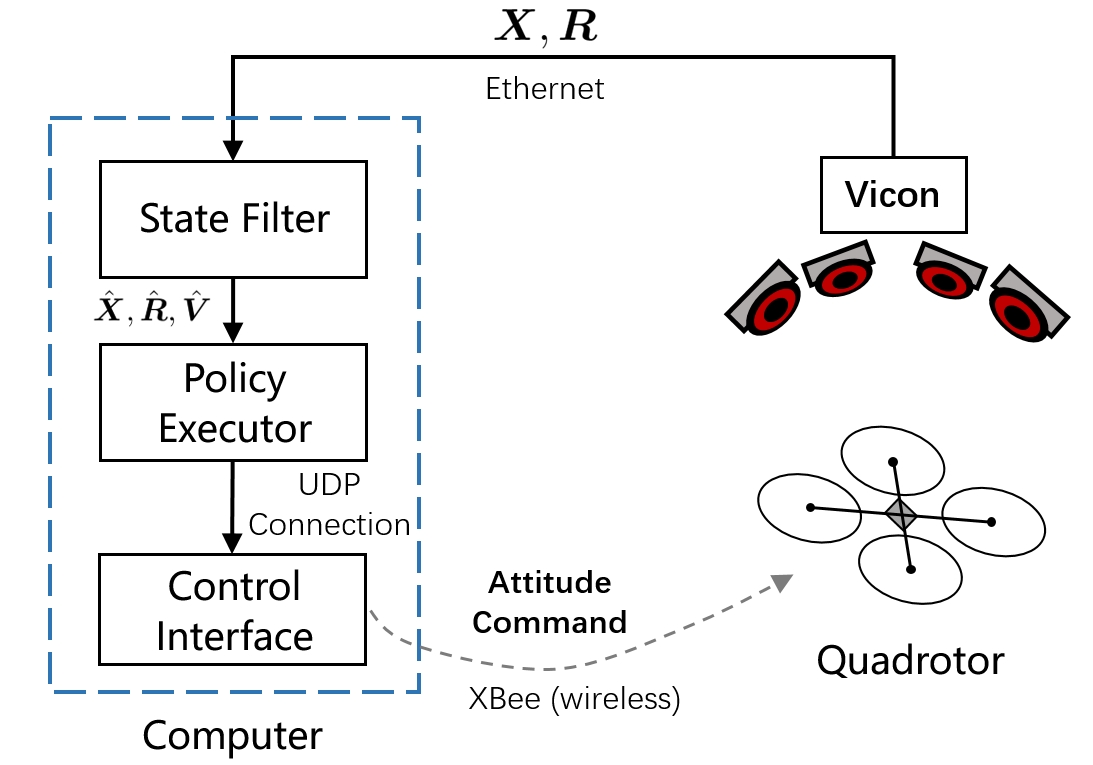}
	\caption{System diagram of our learning-based aggressive flight experiment. $\boldsymbol{X}, \boldsymbol{R}$ are position and orientation from vicon, and $\hat{\boldsymbol{X}}, \hat{\boldsymbol{R}}, \hat{\boldsymbol{V}}$ are filtered position, orientation and velocity.}
	\label{experiment_diag}
	\vspace{-5pt}
\end{figure}

	\begin{table}[!ht]
	\setlength{\abovecaptionskip}{0pt}
	\setlength{\belowcaptionskip}{10pt}
	\captionsetup{justification=centering}
	\caption{Parameters of the quadrotor in our experiments.}
	\label{table_param}
	\begin{center}
		{		
			\begin{tabular}{m{100pt}<{\centering}m{120pt}<{\centering}}
				\hline\hline
				\specialrule{0em}{0pt}{2pt}
				Parameters & Value (units) \\
				\hline
				\specialrule{0em}{0pt}{4pt}
				Weight & 0.547 (kg) \\
				Size & $0.44\times0.44\times0.12$ (m) \\
				Rotor diameter & 20.32 (cm) \\
				Maximum speed & 15 (m/s) \\
				Maximum thrust & 20 (N) \\
				Power (each motors) & 80 (W) \\
				Inertia $I_{xx}, I_{yy}, I_{zz}$ & 0.033, 0.033, 0.058 (kg$\cdot m^2$)\\			\specialrule{0em}{0pt}{2pt}
				\hline\hline
		\end{tabular}}
	\end{center}
	\label{parameters}
	\vspace{-5pt}
\end{table}


	\subsection{Experiments on a Real Quadrotor}
	{In this section, we perform aggressive flight missions in real-world environments using  the policy learned from simulations directly, which also demonstrates the superiority of our method in transferability.	
 We use an Asctec Hummingbird$\footnote{http://www.asctec.de}$ quadrotor to conduct real-world aggressive flight missions, and its specific parameters are shown in Table \ref{parameters}.} The quadrotor is equipped with a wireless communication module and onboard flight controller. In addition, a Vicon motion capture system$\footnote{https://www.vicon.com/}$ is used for obtaining state observation. 
	The configuration of the experiments is shown in Figure  \ref{experiment_diag}. A state filter receives the state information from the Vicon system and sends the filtered states to the policy executor. The quadrotor receives the control signal coming from a control interface and the control interface is connected with the policy executor with user datagram protocol (UDP). It takes less than 1ms for the policy executor to generate an attitude command and the system is updated with a frequency of 100 Hz.

	\begin{figure}[t]
	\vspace{-5pt}
	\setlength{\abovecaptionskip}{0.2cm}
	\setlength{\belowcaptionskip}{-0.cm}
	\centering
	\includegraphics[scale=0.4]{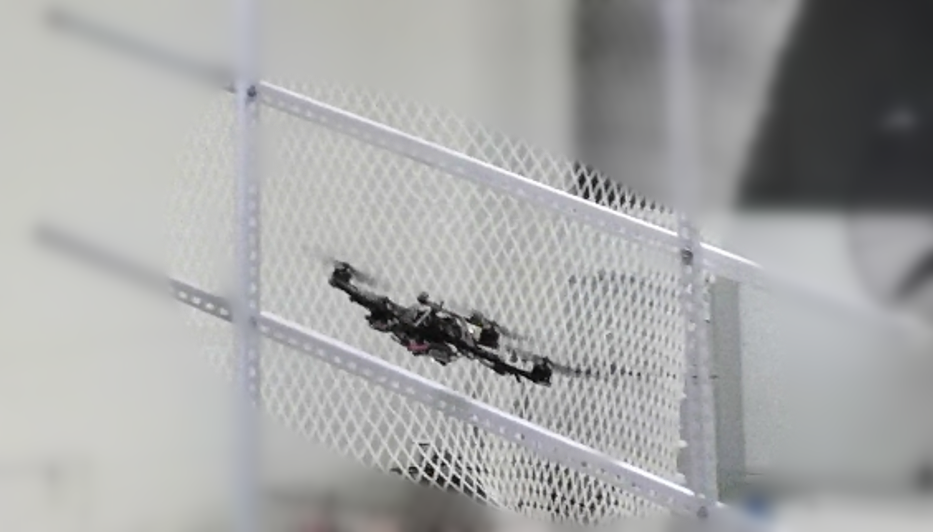}
	\caption{A snapshot of the quadrotor during our aggressive flight experiment.}
	\label{snapshot}
	\vspace{-10pt}
	\end{figure}

	\begin{figure}[t]
	\setlength{\abovecaptionskip}{0.cm}
	\setlength{\belowcaptionskip}{-0.cm}
	\centering
	\subfigure[{The experimental scenes and results.}]{\includegraphics[trim=15 0 0 0, clip, scale=0.33]{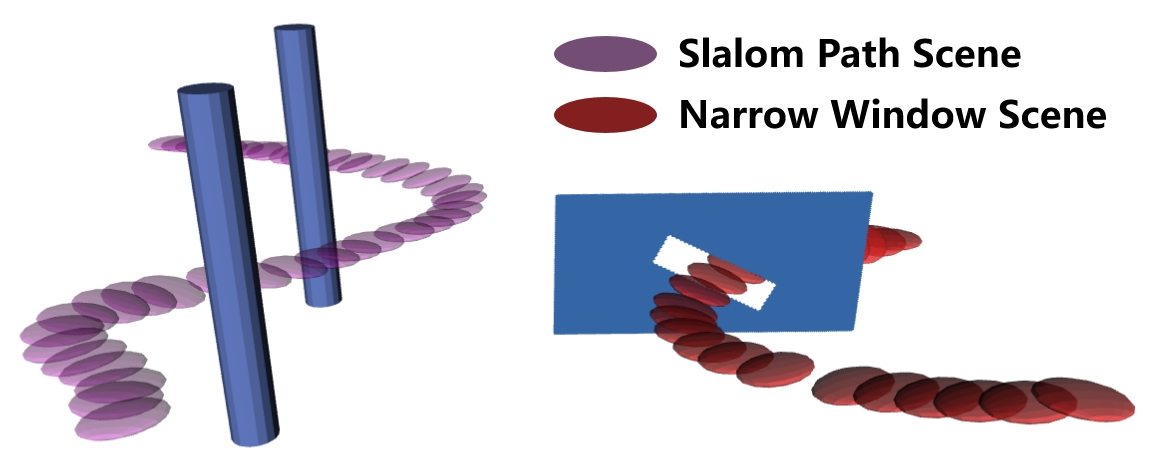}
		\label{3dTraj}} \quad 
	\subfigure[{Comparisons between experimental and simulated trajectories.}]{\includegraphics[trim=10 0 0 0, clip, scale=0.27]{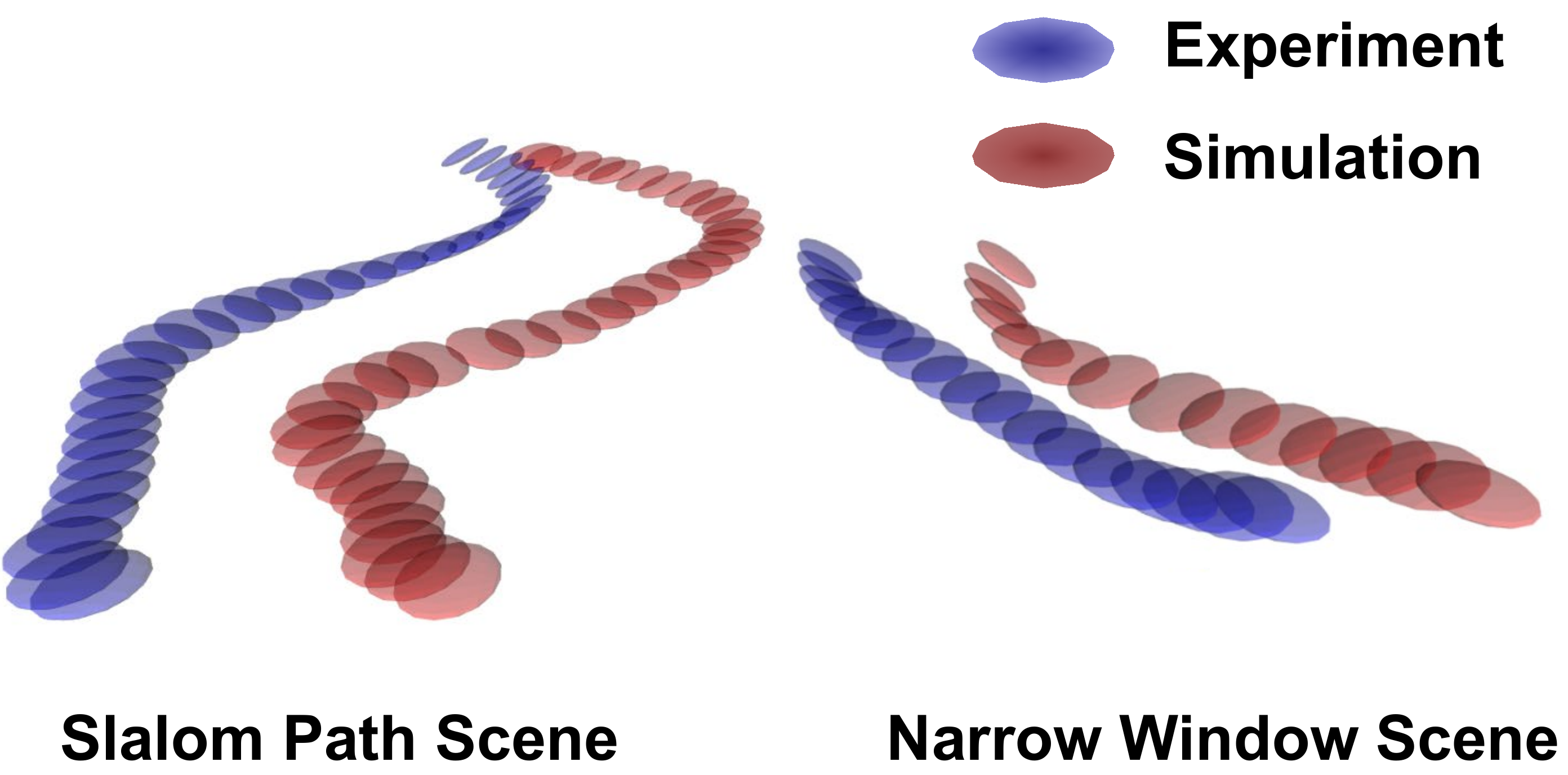}
		\label{3dTrajComp}}  
	\caption{ {The trajectories in our aggressive flight experiments. }}
	
	\vspace{-5pt}
\end{figure}


	\begin{table*}[t]
	
	\scriptsize
	
	\centering
	\captionsetup{justification=centering}
	\caption{{Comparison and ablation experiments in real-world aggressive flight missions.}}
	
	\label{TabExperiment}
	\resizebox{\textwidth}{!}{
		\renewcommand{\multirowsetup}{\centering}
		\begin{threeparttable}
			{
				\begin{tabular}{m{40pt}<{\centering}m{100pt}<{\centering}m{40pt}<{\centering}m{40pt}<{\centering}m{45pt}<{\centering}m{50pt}<{\centering}m{40pt}<{\centering}}					
					\toprule
					Scene &Method & Goal Pos. Error (m)  &  Goal Ang. Error (deg)	&  Sim2real Pos. Difference (m)  &   Sim2real Ang. Difference (deg)  &  Planning Time (s) \\\specialrule{0em}{0pt}{1pt}
					\hline\specialrule{0em}{0pt}{1pt}
					\multirow{6}{*}{\tabincell{c}{\\Narrow \\\specialrule{0em}{1pt}{1pt} Window}} &\textbf{Our Method} &	\textbf{0.034}   & \textbf{2.15} &  \textbf{0.273} &  \textbf{4.78} & \textbf{/} \\\specialrule{0em}{0pt}{1pt}\cline{2-7}\specialrule{0em}{0pt}{1pt}
					&TD3~\cite{TD3} + Curiosity       &	0.055	& 3.34 &  0.448 &  5.50  &  /  \\\specialrule{0em}{0pt}{1pt}\cline{2-7}\specialrule{0em}{0pt}{1pt}
					&TD3~\cite{TD3} + BSE   &	 0.116	& 5.92 &  0.319 &  4.97 &  /\\\specialrule{0em}{0pt}{1pt}\cline{2-7}\specialrule{0em}{0pt}{1pt}
					&TD3~\cite{TD3}   &	 0.521	& 7.51 &  0.484 &  6.76 &  /\\\specialrule{0em}{0pt}{1pt}\cline{2-7}\specialrule{0em}{0pt}{1pt}
					&Nonlinear Tracking Controller~\cite{KumarAggressive}   &   0.049	& 2.82 &  -- &   -- & 41.90 \\\specialrule{0em}{0pt}{1pt}\hline\specialrule{0em}{0pt}{1pt}
					\multirow{4}{*}{\tabincell{c}{\\\\\\Slalom Path}} &	\textbf{Our Method}   & \textbf{0.044} &  \textbf{2.06} &  \textbf{0.187} &  \textbf{6.02} & \textbf{/}	\\\specialrule{0em}{0pt}{1pt}\cline{2-7}\specialrule{0em}{0pt}{1pt}
					&TD3~\cite{TD3} + Curiosity  &	0.068	& 2.76 &  0.371 &  8.39  &  /\\\specialrule{0em}{0pt}{1pt}\cline{2-7}\specialrule{0em}{0pt}{1pt}
					&TD3~\cite{TD3} + BSE   &	 0.084  	& 4.16 &  0.210 &   6.41 &  / \\\specialrule{0em}{0pt}{1pt}\cline{2-7}\specialrule{0em}{0pt}{1pt}
					&TD3~\cite{TD3}   &	 0.456	& 5.31 &  0.431 &  8.81 &  /\\\specialrule{0em}{0pt}{1pt}\cline{2-7}\specialrule{0em}{0pt}{1pt}
					&Nonlinear Tracking Controller~\cite{KumarAggressive}   & 0.057  	& 3.17 &  -- &   -- &  21.47  \\\specialrule{0em}{0pt}{0pt}
					\bottomrule					
					%
				\end{tabular}
				\begin{tablenotes}
					\footnotesize
					\item[1] Reinforcement learning based methods do not require trajectory planning procedure and the planning time is presented as ``/".
					\item[2] The sim2real transferability is tested between the reinforcement learning based  methods, therefore, the simulated error of nonlinear tracking controller is presented as ``--".
			\end{tablenotes}}
		\end{threeparttable}
	}
\end{table*}

\begin{figure*}[t]
	\setlength{\abovecaptionskip}{0.cm}
	\setlength{\belowcaptionskip}{-0.cm}
	\centering
	\subfigure[{Slalom path mission.}]{\includegraphics[trim=10 0 0 10, clip, scale=0.57]{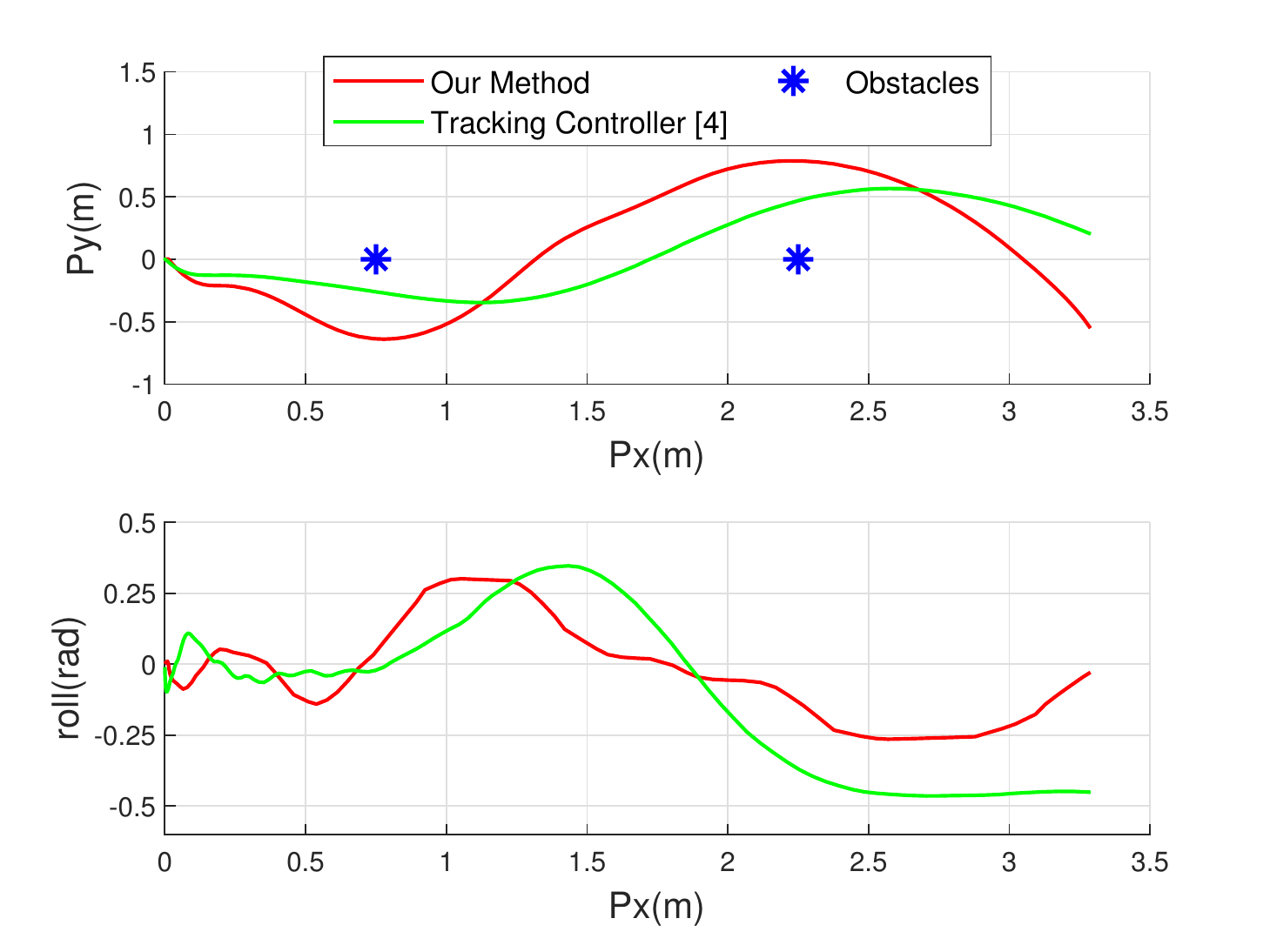}} \quad 
	\subfigure[{Narrow window mission.}]{\includegraphics[trim=10 2 0 10 , clip, scale=0.57]{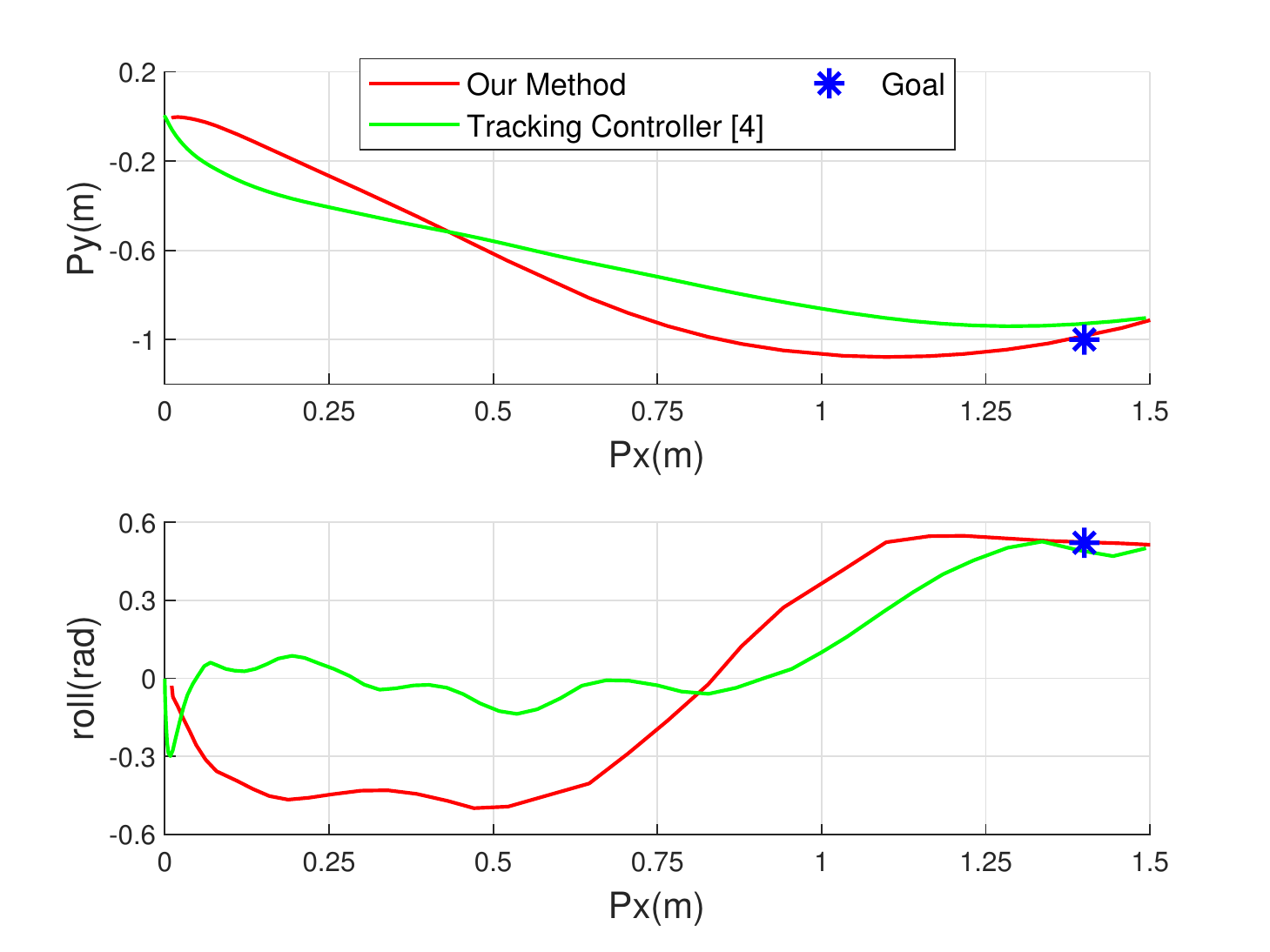}}
	\caption{{Comparisons of position and rotation trajectories in different aggressive flight missions.}}
	\label{ExpPlot}
\end{figure*}

	{We perform aggressive flights in real-world experiments  in both slalom path and narrow window scenes. For the slalom path scene, we used two columnar obstacles placed 1.5m away from each other. For the narrow window scene, we build a narrow window gap with a 30$^\circ$ rotation angle and Figure  \ref{snapshot} is a snapshot of the experiment.
	A 3D illustration of experimental trajectories is shown in Figure  \ref{3dTraj}, and the comparisons between the experimental and simulated trajectories are also shown in Figure  \ref{3dTrajComp}. 
	Because there are differences between the simulations and real-world experiments, 
	the trajectories of the quadrotor in the real-world experiments inevitably deviate from their trajectories in the simulations, as shown in Figure \ref{3dTrajComp}. Despite this, the policy trained with our exploration strategy has a satisfactory transferability and can give a relatively satisfactory action selection in the new states. } 
	
	We conduct ablation and comparison experiments to demonstrate the performance and transferability of our method. In Table \ref{TabExperiment}, we calculate the average position and rotation differences between our trajectories and the goal trajectories (\textbf{Goal Pos. Error} and \textbf{Goal Ang. Error}) to evaluate the performance.  The transferability is evaluated by calculating the differences between the simulated and experimental trajectory results (\textbf{Sim2real Pos. Difference} and \textbf{Sim2real Ang. Difference}).
	To prove the effectiveness of the modules used in our method, we compare our method with the original TD3~\cite{TD3}, TD3 + Curiosity and TD3 + BSE. The results show that both of the curiosity module and the BSE strategy  narrow the differences between the simulated and experimental trajectories, and the performance is improved as well.

	{To demonstrate the performance of our curiosity-driven reinforcement learning method further, we compare our method with the planning-based aggressive flight method~\cite{KumarAggressive}. The method  uses the aggressive trajectory planning~\cite{LiuTrajPlan} and a nonlinear controller~\cite{LeeCtrler}. The quantitative results are shown in  Table \ref{TabExperiment} and the 
	trajectories are plotted in Figure \ref{ExpPlot}. The  results in Table \ref{TabExperiment} show that our method obtains a smaller error without trajectory planning, which demonstrate that our method have a better performance and transferability when compared to~\cite{LiuTrajPlan}. In Figure \ref{ExpPlot}, our method is more far from the obstacles in the slalom path mission and more closer to the goal in narrow window mission.
}

	

	\section{CONCLUSIONS}
	{In this paper, we confront the problem of aggressive flight with a curiosity-driven reinforcement learning method. 	
	We introduce a similarity-based curiosity module to overcome the  sparse reward problem in reinforcement learning, which can achieve a satisfactory performance with a fast convergence speed. Besides, the BSE strategy improves the robustness of our learning-based controller, which makes the policy trained in simulation can be directly used with a real quadrotor. Our method accelerates the execution of the aggressive movements by reducing the dependence of trajectory  planning step, and it shows the sim2real transferability.}
	
	
{In the future, we will try to conduct
more aggressive tasks with higher speed and rotation angles through reinforcement learning. Since
the higher rotation angle leads to higher traversing speed, the quadrotor needs faster execution rate and
more precise control commands. To handle these, we plan to deploy our algorithm onboard with the matrix arithmetic method. Besides,  model-based reinforcement learning methods will be considered.
The dynamic model of the quadrotor will be used as a state predictor to make full use of the
theoretical information in the learning process and control the quadrotor more precisely.
We will also explore the possibilities of visual-based reinforcement learning methods for aggressive flight, in which different cameras can be used, such as the event-based and RGB-D cameras.}

	\bibliographystyle{IEEEtranTIE}
	\bibliography{TXT_21-TIE-0617}

\section{Appendix}
\subsection{Experiments in unstructured environments}
\begin{figure}[H]
	\centering
	\subfigure[{Scene 1}]{\includegraphics[scale=0.5]{unstructered1}} \quad 
	\subfigure[{Scene 2}]{\includegraphics[scale=0.7]{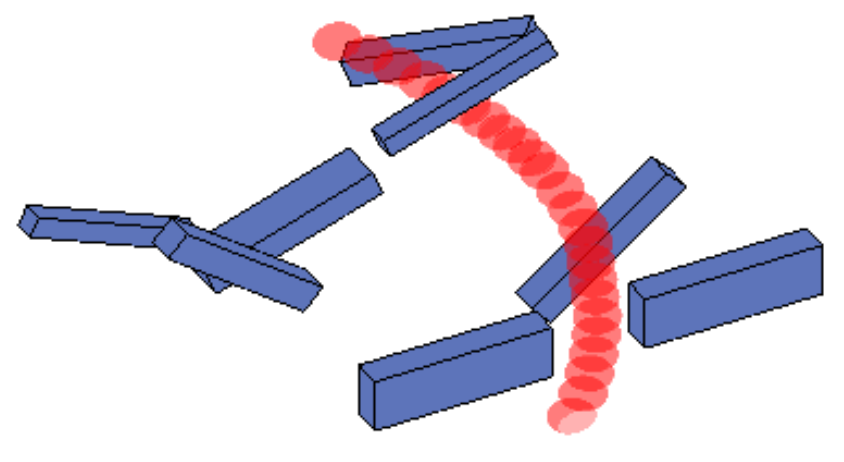}} \quad 
	\subfigure[{Scene 3}]{\includegraphics[scale=0.7]{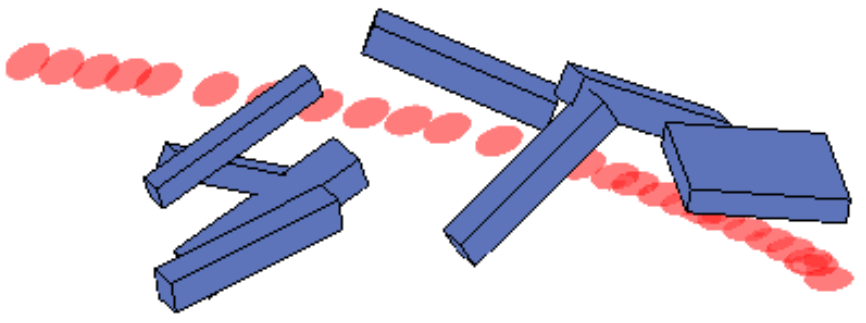}}
	\caption{The trajectories in an unstructured environment.}
	\label{unstructered3}
\end{figure}

We add some experiments in unstructured environments by running Monte-Carlo simulations for a collection of scenes containing random unstructured obstacles. In the generated scenes, there exists at least one path that is in accord with quadrotor's dynamics constraints for the agent to traverse across.  The result demonstrates that our method not only performs well in classical aggressive flight missions like specific narrow window and slalom path scenarios, but also works well in other unstructured environments, as shown in Figure \ref{unstructered3}.
To show the 3d geometry of the unstructured environments in a clearer way, we have uploaded the experimental video for the navigation in an unstructured environment and you can watch the video at
\underline{\textcolor{blue}{https://youtu.be/5ZI9qyZNwjo}}.

\subsection{Scene with two narrow windows}

\begin{figure}[H]
	\centering
	\subfigure[viewpoint 1]{\includegraphics[scale=0.3]{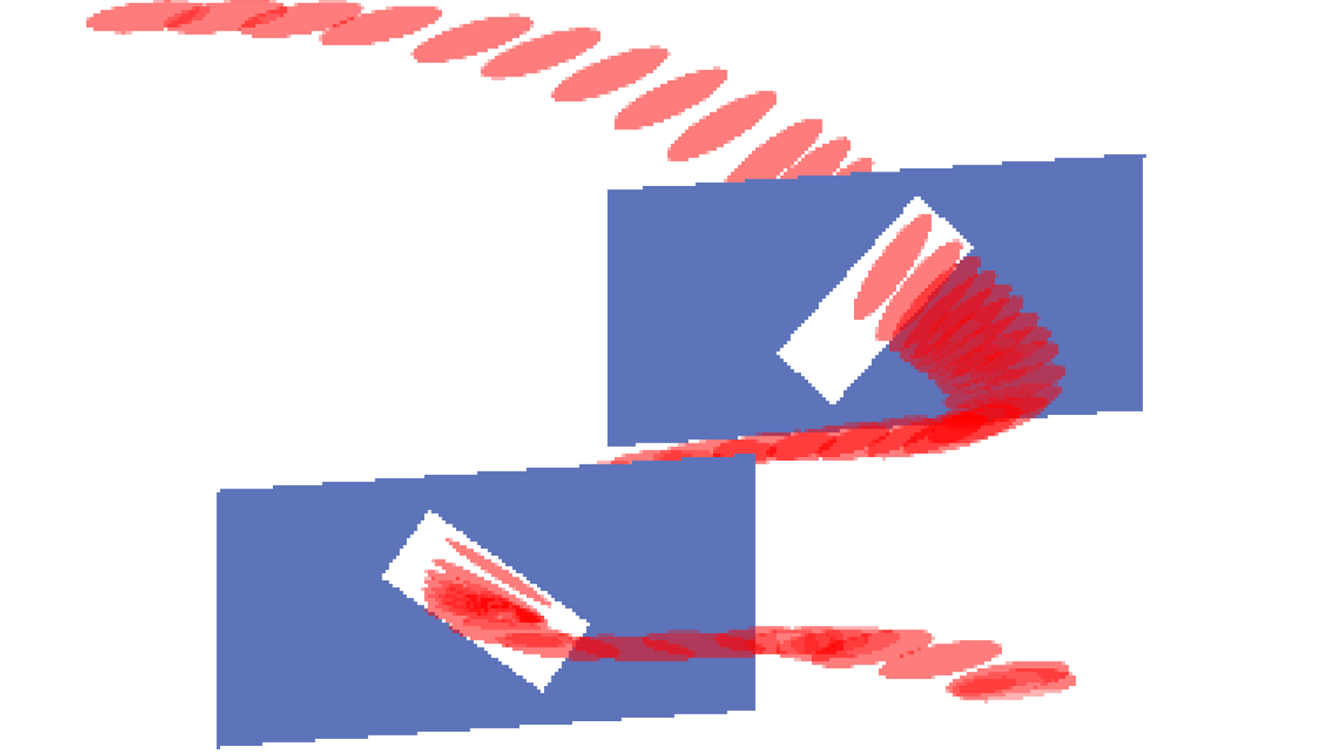}} \quad 
	\subfigure[viewpoint 2]{\includegraphics[scale=0.25]{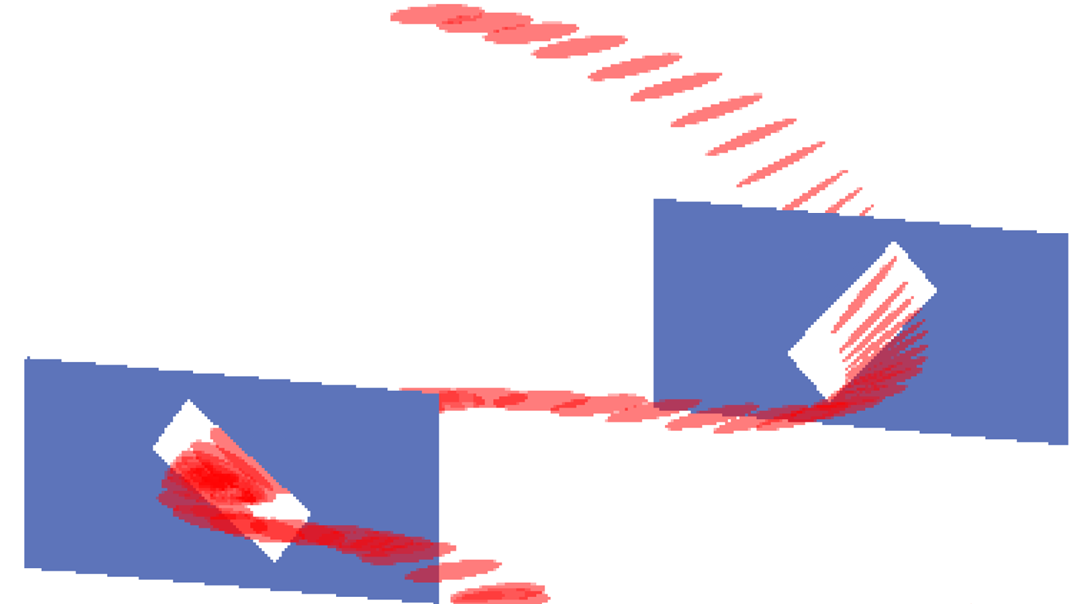}} 
	\caption{Scene with two narrow windows.}
	\label{viewpoint}
\end{figure}

The simulation experiment with two narrow windows is also conducted  to prove that our method works well in more complex environment, as shown in Figure \ref{viewpoint}.

\subsection{Visual comparison of our work and [4]}

\begin{figure}[H]
	\setlength{\belowcaptionskip}{-0.5cm}
	\centering
	\subfigure[{Slalom path mission.}]{\includegraphics[scale=0.35]{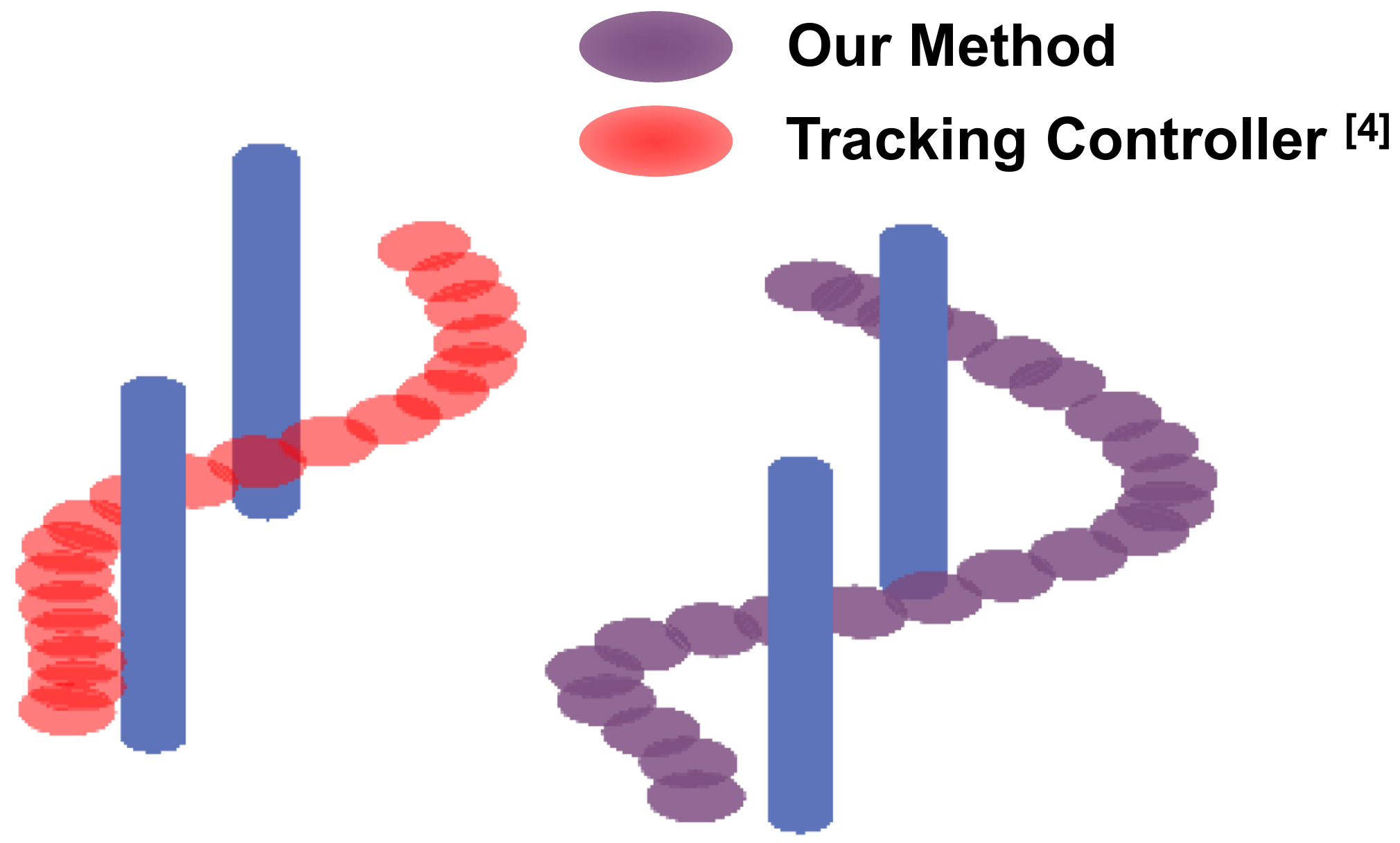}} \quad 
	\subfigure[{Narrow window mission.}]{\includegraphics[scale=0.35]{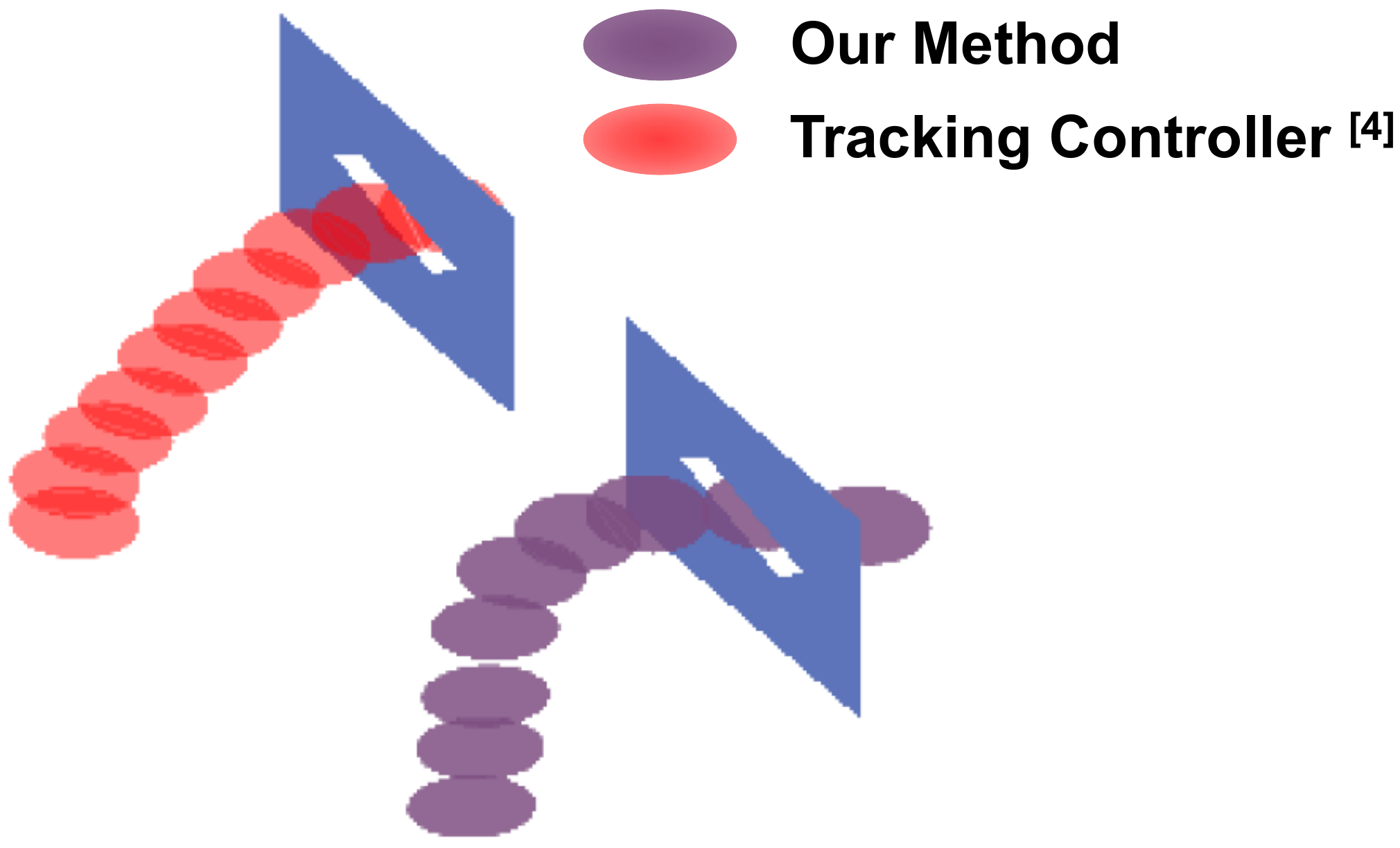}}
	\caption{Visual comparison of different methods.}
	\label{Visual}
\end{figure}

We plot the trajectories of different methods in Table IV for a clearer visual comparison. As shown in Figure \ref{Visual}, our method is more far from
the obstacles in the slalom path mission and has a smoother curve that is in accord with the dynamics constraints of the quadrotor. We can also find  that our method is more closer to
the goal in narrow window mission in Figure 13, compared to traditional tracking controller [4].

\begin{table*}[t]
	\caption{The comparison of our method and state-of-the-art methods.}
	\begin{center}		
		
		\begin{tabular}{|c|c|c|c|c|c|}
			\hline
			\multirow{2}{*}{\textbf{Method}} &\multicolumn{2}{c|}{\textbf{{Trajectory Planning}}}& \multicolumn{2}{c|}{\textbf{{Pre-training}}}&	\multirow{2}{*}{\textbf{Applications}} \\										
			\cline{2-5}
			
			&Time&Frequency&Time&Frequency&\\
			
			\hline 
			Loianno \textit{et al.} [4] & Dozens of seconds& Before every flight& \multicolumn{2}{c|}{{Not necessary }} & Aggressive flights\\					
			\hline					
			Liu \textit{et al.} [6]  & Dozens of seconds& Before every flight& \multicolumn{2}{c|}{{Not necessary }} & Aggressive flights\\					
			\hline
			Falanga \textit{et al.} [21] &\multicolumn{2}{c|}{{Real-time}}&\multicolumn{2}{c|}{{Not necessary }} &Obstacle avoidance\\					
			\hline
			Ryll \textit{et al.} [22] &\multicolumn{2}{c|}{{Real-time}}&\multicolumn{2}{c|}{{Not necessary }} &high speed flights\\					
			\hline					
			Ours & \multicolumn{2}{c|}{{Not necessary }}& 1 hour & Once & Aggressive flights\\					
			\hline

		\end{tabular}
	\end{center}
	\label{tab:transfer}
\end{table*} 

\subsection{Comparison of different methods}

We add a table to compare our method with some other state-of-the-art methods in terms of various aspects.

From Table \ref{tab:transfer}, we can find that our method can execute real-time aggressive flight missions without trajectory planning after the one-time offline pre-training, while traditional methods [4], [6] need dozens of seconds for trajectory planning before every flight.  Though reference [21] can conduct trajectory planning in real-time and [22] is able to perform efficient trajectory planning for high linear speed flight, our method concentrates on aggressive flight, in which the agents fly with both high linear and angular speed.

\subsection{Generalization of our method}

\begin{figure}[htb!]
	\centering
	\subfigure[{Position trajectories.}]{\includegraphics[scale=0.6]{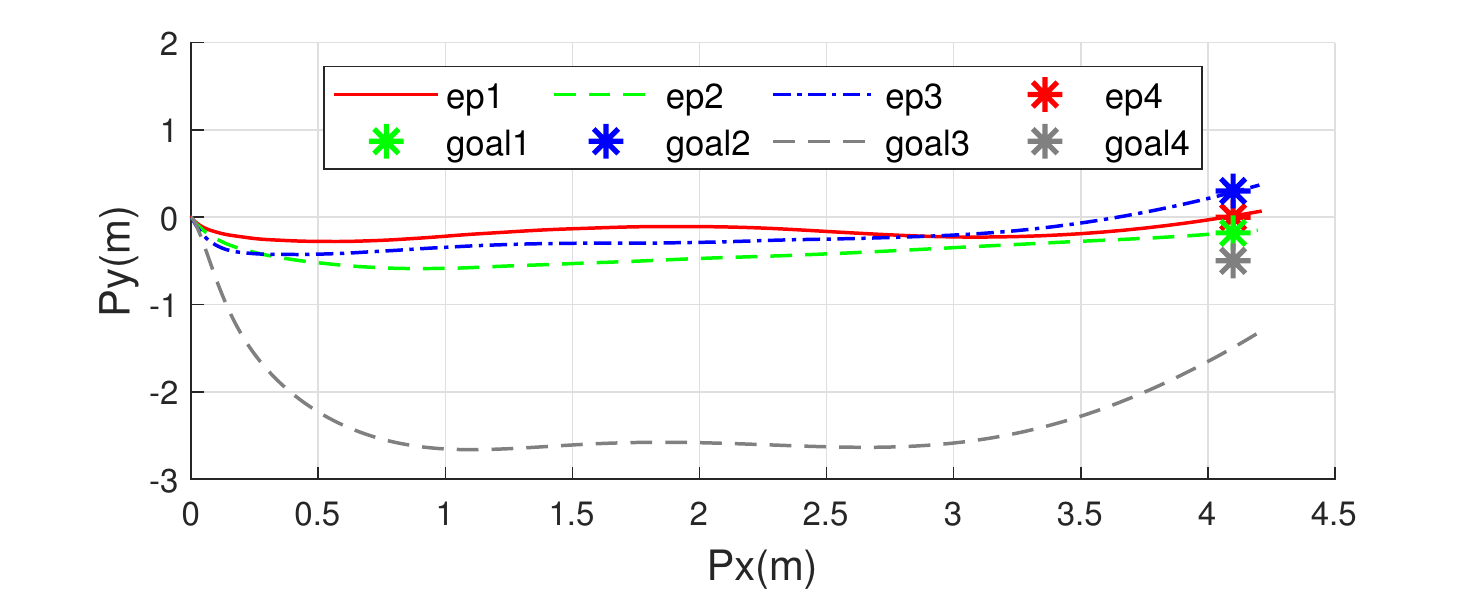}} \quad 
	\subfigure[{Rotation trajectories.}]{\includegraphics[scale=0.6]{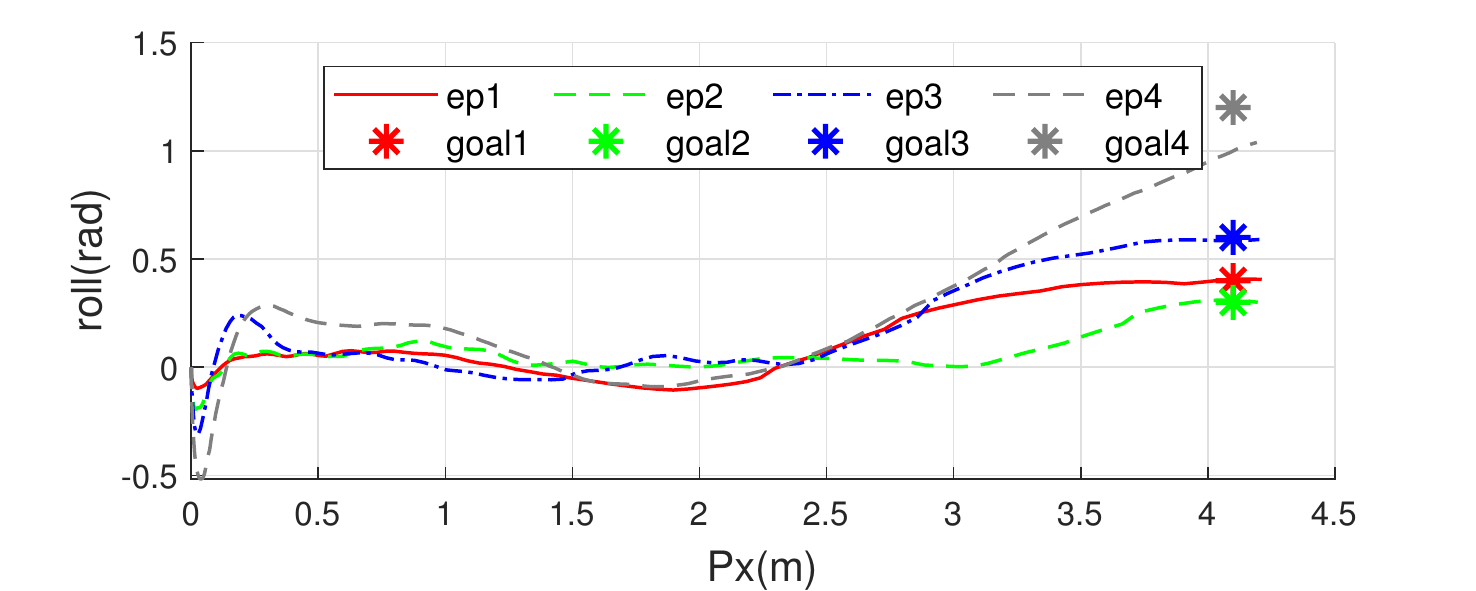}} 
	\caption{Position and rotation trajectories in narrow window scene.}
	\label{fail}
\end{figure}

We perform the experiments on diverse narrow windows
using the same pre-trained policy to demonstrate the
generalization  of our method in Figure 8. The experiments shown in Figure 8 are the successful cases, and we continue to  change the setup of the narrow window  to explore when the negative results appears. We find that when the rotation angle and the distance are set nearly to  1.2 rad and -0.5 m, the unsuccessful cases begin to  appear, as shown with gray line in Figure \ref{fail}.
Based on the experimental results, we find that when marked changes appear in environment, our method may fail to fulfill the task. How to further improve the generalization of our method is one of our further research directions.

\end{document}